\documentclass[letterpaper, 10 pt, conference]{ieeeconf}
\newif\ifarxiv \arxivtrue
\newif\ifzh \zhfalse
\usepackage{latexml}
\IEEEoverridecommandlockouts
\usepackage{amsmath,amssymb}
\usepackage{graphicx}
\usepackage{booktabs,multirow,array}
\usepackage{xcolor}
\usepackage{tikz}
\usetikzlibrary{arrows.meta,positioning,calc,fit,backgrounds,shapes.geometric}
\usepackage{pgfplots}
\usepackage{pgfplotstable}
\makeatletter
\@ifundefined{ifarxiv}{\newif\ifarxiv \arxivfalse}{}
\makeatother
\newcommand{\supp}[2]{\ifarxiv#1\else#2\fi}
\pgfplotsset{compat=1.18}
\usepgfplotslibrary{groupplots,fillbetween}
\pgfplotsset{discard if not/.style 2 args={x filter/.append code={\edef\tempa{\thisrow{#1}}\edef\tempb{#2}\ifx\tempa\tempb\else\fi}}}
\makeatletter
\def\caption@documentclass{standard}
\makeatother
\usepackage{subcaption}
\usepackage{url}
\ifarxiv\usepackage[hidelinks]{hyperref}\else\usepackage[hidelinks,draft]{hyperref}\fi
\newcommand{\submissionpagestyle}{%
  \ifarxiv\pagestyle{plain}\thispagestyle{plain}%
  \else\pagestyle{empty}\thispagestyle{empty}\fi}
\usepackage{cite}
\usepackage{stfloats}
\usepackage{placeins}
\usepackage{listings}
\lstdefinestyle{papercode}{
  columns=fullflexible,keepspaces=true,showstringspaces=false,
  basicstyle=\ttfamily\color{codeInk},
  keywordstyle=\bfseries\color{codeKey},
  stringstyle=\color{codeStr},
  commentstyle=\itshape\color{codeComment},
  identifierstyle=,numberstyle=,
  emph={[2]True,False,None},emphstyle={[2]\color{codeConst}},
  rulecolor=\color{ruleGray},framerule=.35pt,framesep=3pt}
\lstdefinestyle{agentexcerpt}{style=papercode,
  basicstyle=\ttfamily\fontsize{7.5}{8.5}\selectfont\color{codeInk},
  breaklines=true,breakatwhitespace=false,breakindent=8pt,
  aboveskip=3pt,belowskip=3pt,frame=single}
\lstdefinestyle{apilisting}{style=papercode,language=Python,
  morekeywords={True,False,None},
  emph={[3]make_gym,get_demos,ingest,reset,act,step,observe,update,
        close,Agent},
  emphstyle={[3]\color{codeFunc}},
  basicstyle=\ttfamily\scriptsize\color{codeInk},
  aboveskip=5pt,belowskip=0pt,frame=single,
  xleftmargin=5.5pt,framexleftmargin=2pt}
\iflatexml
\newcommand{\agentfiletext}{boxtext}
\newcommand{\agentfile}[4][boxPrompt]{%
  \par\noindent\textbf{#3}\par
  \def\agentfilestyle{#2}%
  \ifx\agentfilestyle\agentfiletext
    \lstinputlisting[basicstyle=\ttfamily\footnotesize,columns=fullflexible,keepspaces=true,showstringspaces=false,upquote=true]{html/#4}%
  \else
    \lstinputlisting[style=papercode,language=Python,basicstyle=\ttfamily\footnotesize\color{codeInk},breaklines=true,upquote=true]{#4}%
  \fi
}
\else
\usepackage[many]{tcolorbox}
\tcbuselibrary{listings,breakable,skins}
\newcommand{\codefont}{\fontencoding{T1}\renewcommand{\ttdefault}{zi4}\ttfamily}
\definecolor{boxPrompt}{HTML}{005B8E}
\definecolor{boxDoc}{HTML}{006F51}
\definecolor{boxCode}{HTML}{3C3C3C}
\tcbset{paperbox/.style={enhanced, breakable, arc=1.2mm, boxrule=0.6pt,
  colframe=#1, colbacktitle=#1, coltitle=white, colback=gray!4!white,
  fonttitle=\bfseries\sffamily\footnotesize, toptitle=0.6mm, bottomtitle=0.6mm,
  left=2.5mm, right=2.5mm, top=1.5mm, bottom=1.5mm}}
\lstdefinestyle{boxtext}{columns=fullflexible,keepspaces=true,
  basicstyle=\codefont\fontsize{7.1}{8.6}\selectfont\color{black},
  breaklines=true,breakatwhitespace=true,breakindent=1em,
  showstringspaces=false,upquote=true,literate={*}{{\char42}}1,
  aboveskip=0pt,belowskip=0pt}
\lstdefinestyle{boxcode}{style=papercode,language=Python,
  basicstyle=\codefont\fontsize{7.1}{8.6}\selectfont\color{black},
  breaklines=true,breakindent=16pt,upquote=true,literate={*}{{\char42}}1,
  commentstyle=\color{codeComment},aboveskip=0pt,belowskip=0pt}
\newtcbinputlisting{\agentfile}[4][boxPrompt]{paperbox=#1, title={#3},
  listing only, listing file={#4}, listing options={style=#2}}
\fi
\usepackage{microtype}
\microtypesetup{protrusion=false}
\newsavebox{\tablebox}
\graphicspath{{figures/}{figures/tasks/}}
\providecommand{\zhen}[2]{\ifzh #1\else #2\fi}
\newcommand{\benchname}{BiGym~2.0}
\newcommand{\ntasks}{20}

\newcommand{\ndemos}{60}
\newcommand{\neps}{100}

\newcommand{\nseeds}{3}
\newcommand{\seedbase}{620000}

\ifdefined\marksmode
  \newcommand{\todo}[1]{{\color{red}[#1]}}
  
\else
  \newcommand{\todo}[1]{}
  
\fi
\newcommand{\pizero}{$\pi_{0.5}$}

\newcommand{\drq}{DrQ-v2+}
\definecolor{otblue}{HTML}{0072B2}
\definecolor{otorange}{HTML}{E69F00}
\definecolor{otteal}{HTML}{009E73}
\definecolor{otpurple}{HTML}{CC79A7}
\definecolor{otgray}{HTML}{5A5A5A}
\definecolor{otred}{HTML}{D55E00}

\definecolor{okBlue}{HTML}{0072B2}
\definecolor{okGreen}{HTML}{009E73}
\definecolor{okVermillion}{HTML}{D55E00}
\definecolor{okOrange}{HTML}{E69F00}
\definecolor{okSky}{HTML}{56B4E9}
\definecolor{okPurple}{HTML}{CC79A7}
\definecolor{okYellow}{HTML}{F0E442}
\definecolor{okGray}{HTML}{5A5A5A}
\definecolor{inkGray}{HTML}{3C3C3C}
\definecolor{ruleGray}{HTML}{B8B8B8}
\colorlet{cACT}{okBlue}
\colorlet{cDP}{okGreen}
\colorlet{cCQNAS}{okVermillion}
\colorlet{cRef}{okGray}          %
\colorlet{cVLA}{okPurple}        %
\colorlet{cController}{okOrange} %
\colorlet{cData}{okSky}          %
\colorlet{cProtocol}{okPurple}   %
\colorlet{cNeutral}{okGray}
\definecolor{catReach}{HTML}{3F7B8A}
\definecolor{catTable}{HTML}{9A7336}
\definecolor{catDish}{HTML}{55639B}
\definecolor{catCounter}{HTML}{8A5268}
\definecolor{codeInk}{HTML}{3A4449}
\definecolor{codeKey}{HTML}{B4453F}
\definecolor{codeStr}{HTML}{6B7A18}
\definecolor{codeComment}{HTML}{6E7C6E}
\definecolor{codeConst}{HTML}{A8468C}
\definecolor{codeFunc}{HTML}{2E7593}
\pgfplotsset{
  every axis/.append style={
    axis lines=left, axis line style={ruleGray, line width=0.5pt},
    tick style={ruleGray}, tick align=outside, minor tick num=0,
    tick label style={font=\sffamily\tiny, text=inkGray},
    label style={font=\sffamily\tiny, text=inkGray},
    title style={font=\sffamily\scriptsize, text=inkGray},
    legend style={font=\sffamily\tiny, draw=none, fill=none, text=inkGray},
    clip=false,
  },
}
\tikzset{every picture/.append style={font=\sffamily\scriptsize}}

\begin{document}
\iflatexml
\title{\parbox[c]{17pt}{\includegraphics[width=15.5pt,alt={BiGym logo}]{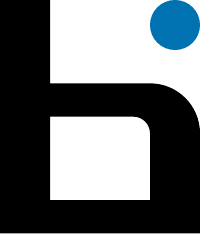}}\hspace{0.28em}\benchname: Benchmarking Learned and Agent-Developed Policies for Humanoid Household Manipulation}
\else
\title{\LARGE \bf \ifarxiv\raisebox{-0.14\height}{\includegraphics[height=0.95em]{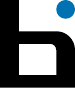}}\hspace{0.28em}\fi\benchname: Benchmarking Learned and Agent-Developed Policies for Humanoid Household Manipulation}
\fi
\iflatexml
\author{{Zexi Zhang$^{*}$, Zecheng Zhu$^{*}$, Zidong Chen, Zulkhuu Tuya and Stephen James}\\
{\small $^{*}$Equal contribution. All authors are with the Department of Computing, Imperial College London, UK.}}
\else
\author{Zexi Zhang$^{*}$, Zecheng Zhu$^{*}$, Zidong Chen, Zulkhuu Tuya and Stephen James%
\thanks{$^{*}$Equal contribution. All authors are with the Department of Computing, Imperial College London, UK.}}
\fi
\bstctlcite{IEEE:BSTcontrol}
\iflatexml
\maketitle
\begin{figure*}
\centering
\includegraphics[width=0.85\textwidth]{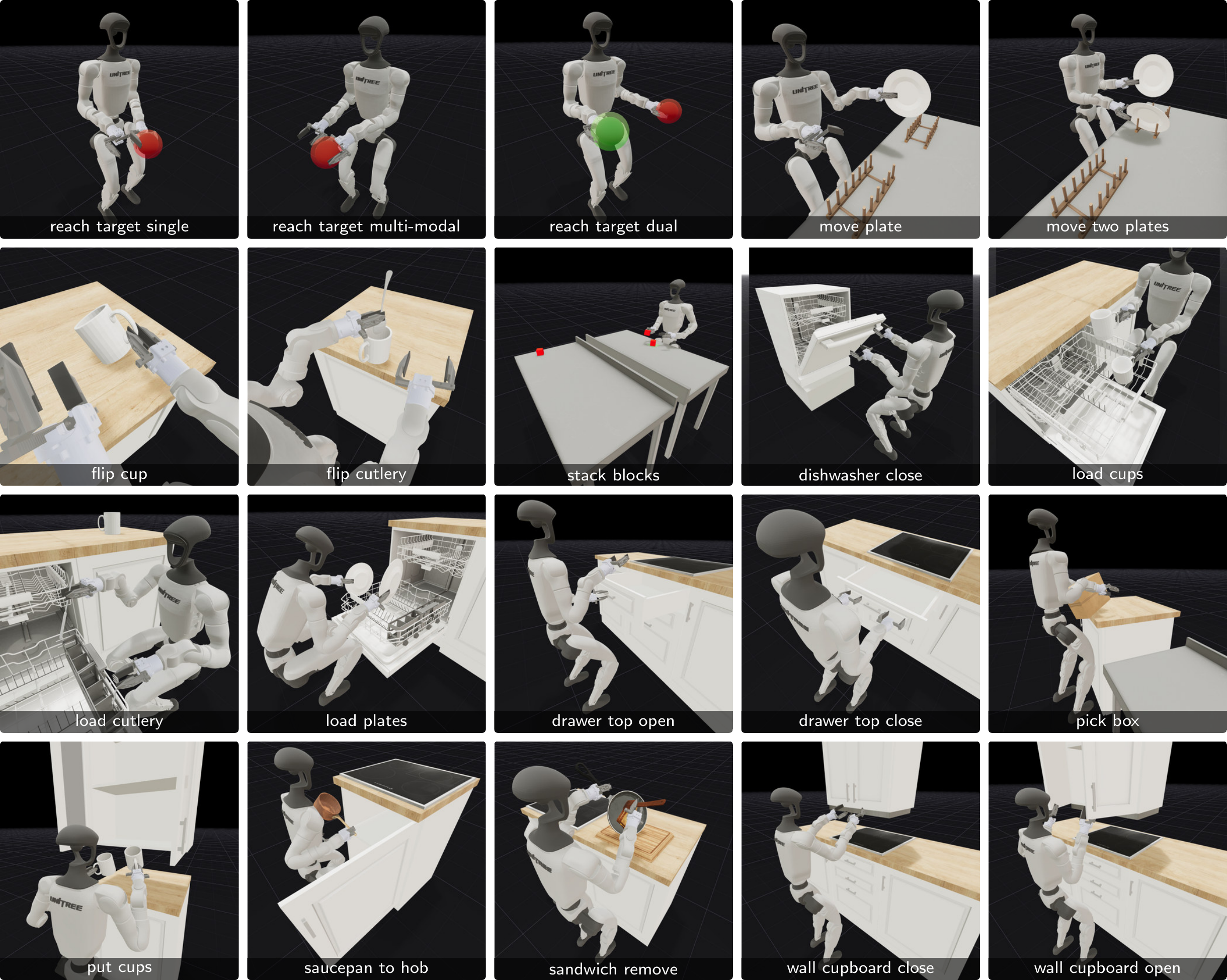}
\caption{\textbf{\benchname{}: Auditing learned and agent-developed household manipulation on a walking humanoid.} The suite evaluates data-driven policies and coding agents across \ntasks{} tasks spanning reaching, tabletop, dishwasher and kitchen-counter scenes. Tasks test hand choice, bimanual coordination, articulated fixtures, multi-object transport and cross-workspace block stacking. Human demonstrations, learned policies and agent programs execute through the same frozen whole-body controller, evaluating manipulation reliability under dynamic balance and footfall disturbances.}
\label{fig:tasks}
\end{figure*}
\else
\IEEEaftertitletext{%
\begin{minipage}{\textwidth}
  \centering
  \resizebox{0.85\textwidth}{!}{\begin{tikzpicture}[x=1cm,y=1cm]
\def\tkcols{5}%
\def\tkgap{0.12}%
\def\tkw{3.444}           %
\def\tkbar{0.32}          %
\newcommand{\tkcell}[3]{%
  \pgfmathtruncatemacro\tkc{mod(#3,\tkcols)}%
  \pgfmathtruncatemacro\tkr{floor(#3/\tkcols)}%
  \pgfmathsetmacro\tkx{\tkc*(\tkw+\tkgap)}%
  \pgfmathsetmacro\tky{-\tkr*(\tkw+\tkgap)}%
  \begin{scope}%
    \clip[rounded corners=1.5pt] (\tkx,\tky) rectangle (\tkx+\tkw,\tky-\tkw);%
    \node[anchor=north west,inner sep=0pt] at (\tkx,\tky)%
      {\includegraphics[width=\tkw cm]{#1}};%
    \fill[black,opacity=.55] (\tkx,\tky-\tkw) rectangle (\tkx+\tkw,\tky-\tkw+\tkbar);%
  \end{scope}%
  \node[font=\sffamily\scriptsize,text=white]%
    at (\tkx+0.5*\tkw,\tky-\tkw+0.5*\tkbar) {#2};}%
\path[use as bounding box] (0,0) rectangle (17.7,-14.136);

\tkcell{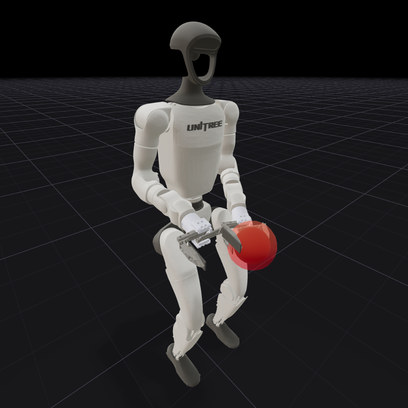}{reach target single}{0}
\tkcell{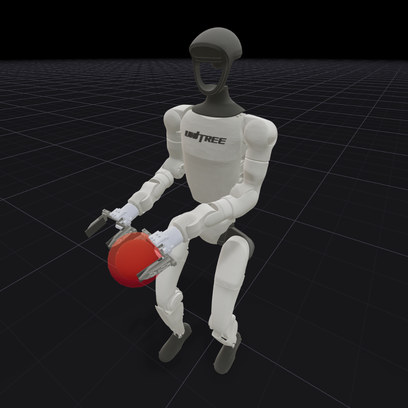}{reach target multi-modal}{1}
\tkcell{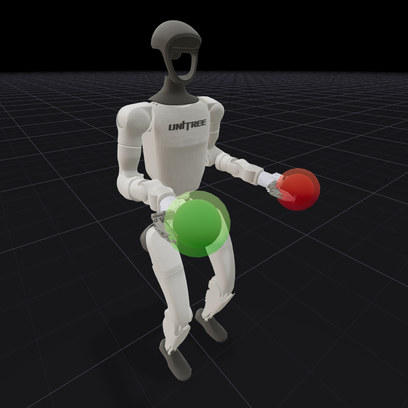}{reach target dual}{2}
\tkcell{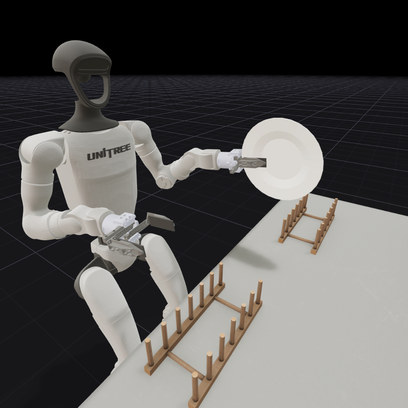}{move plate}{3}
\tkcell{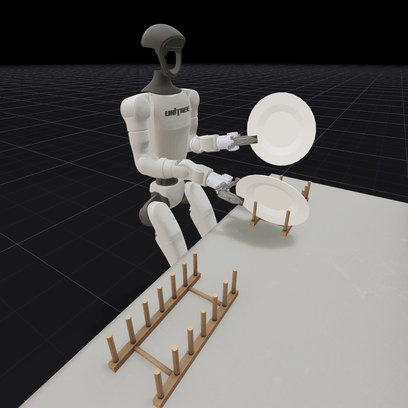}{move two plates}{4}
\tkcell{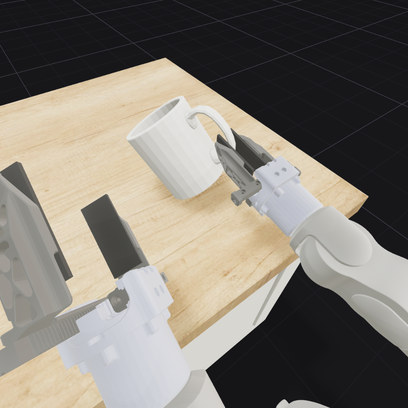}{flip cup}{5}
\tkcell{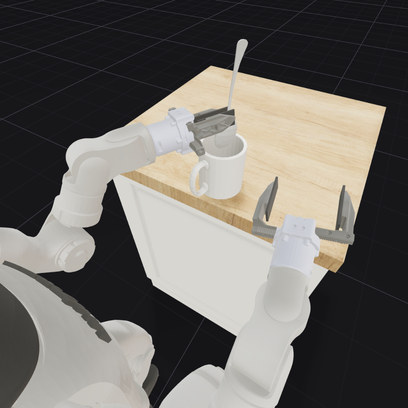}{flip cutlery}{6}
\tkcell{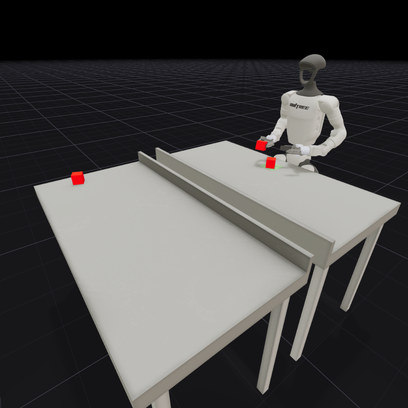}{stack blocks}{7}
\tkcell{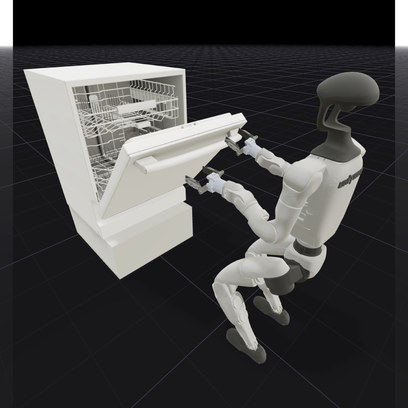}{dishwasher close}{8}
\tkcell{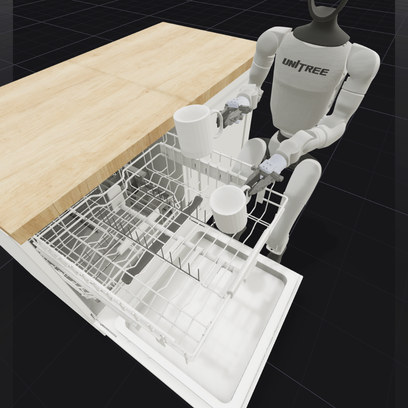}{load cups}{9}
\tkcell{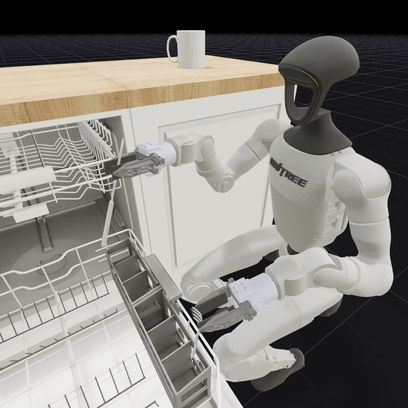}{load cutlery}{10}
\tkcell{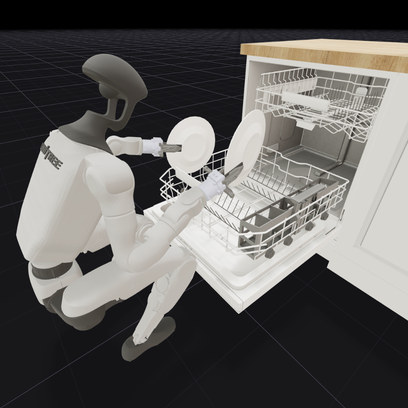}{load plates}{11}
\tkcell{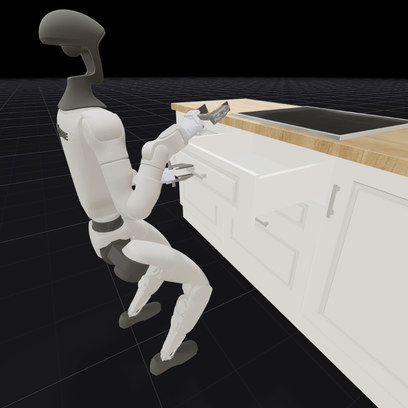}{drawer top open}{12}
\tkcell{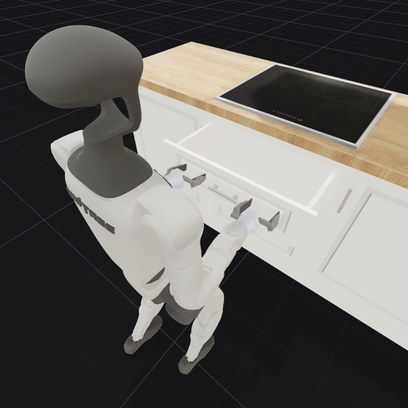}{drawer top close}{13}
\tkcell{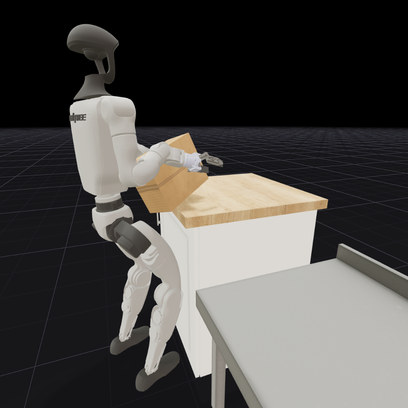}{pick box}{14}
\tkcell{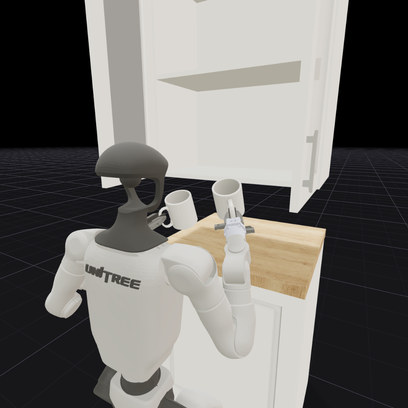}{put cups}{15}
\tkcell{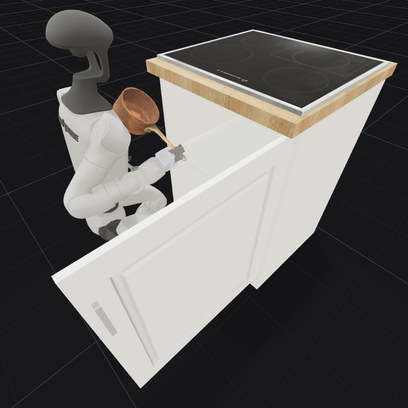}{saucepan to hob}{16}
\tkcell{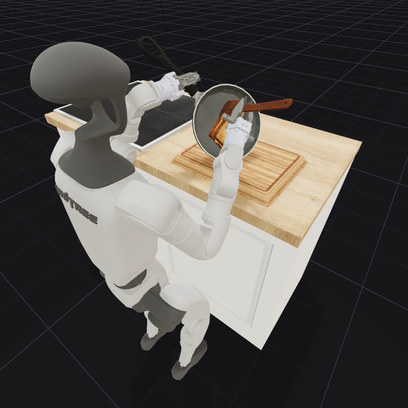}{sandwich remove}{17}
\tkcell{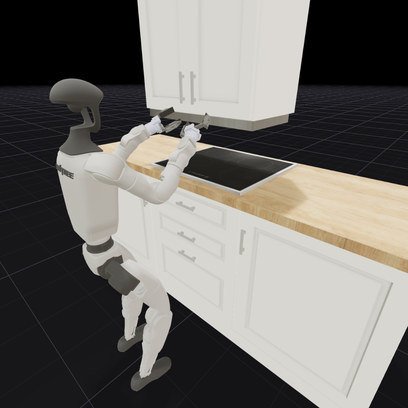}{wall cupboard close}{18}
\tkcell{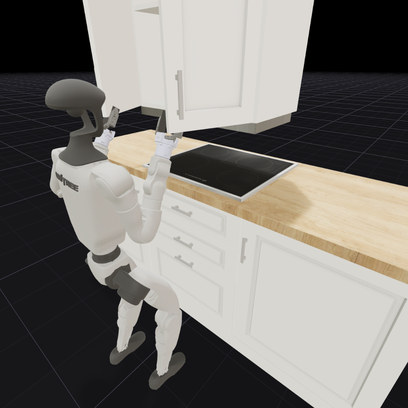}{wall cupboard open}{19}
\end{tikzpicture}%
}
  \captionof{figure}{\textbf{\benchname{}: Auditing learned and agent-developed household manipulation on a walking humanoid.} The suite evaluates data-driven policies and coding agents across \ntasks{} tasks spanning reaching, tabletop, dishwasher and kitchen-counter scenes. Tasks test hand choice, bimanual coordination, articulated fixtures, multi-object transport and cross-workspace block stacking. Human demonstrations, learned policies and agent programs execute through the same frozen whole-body controller, evaluating manipulation reliability under dynamic balance and footfall disturbances.}
  \label{fig:tasks}
\end{minipage}
\vspace{1mm}}
\maketitle
\submissionpagestyle
\fi
\begin{abstract}
Humanoid household manipulation requires the arms to act while the body balances, steps and changes posture. We present \benchname{}, an adaptation of BiGym for the Unitree G1 across \ntasks{} household tasks using a unified whole-body controller for demonstration and evaluation. The suite provides \ndemos{} native human virtual-reality demonstrations per task with synchronised multi-camera views and full-body execution records. We benchmark vision-language-action fine-tuning, imitation learning, demo-driven reinforcement learning, and cold-start coding agents given the interaction budget of online reinforcement learning. With the same onboard views, proprioception and whole-body controller for every method, vision-language-action fine-tuning has the highest nine-task mean, and agent-developed programs outperform every demo-driven reinforcement learning baseline on this mean and lead on bimanual reaching. Cross-workspace stacking remains open, \pizero{} stays low on pick-box, and multi-object transport is hard for imitation learning, demo-driven reinforcement learning and coding agents. All environments, human demonstrations, and evaluation traces are open-sourced\supp{ at \url{https://github.com/swirl-uk/BiGym2}}{}.
\end{abstract}

\begin{figure*}[!b]
  \centering
\iflatexml
  \includegraphics[width=\textwidth]{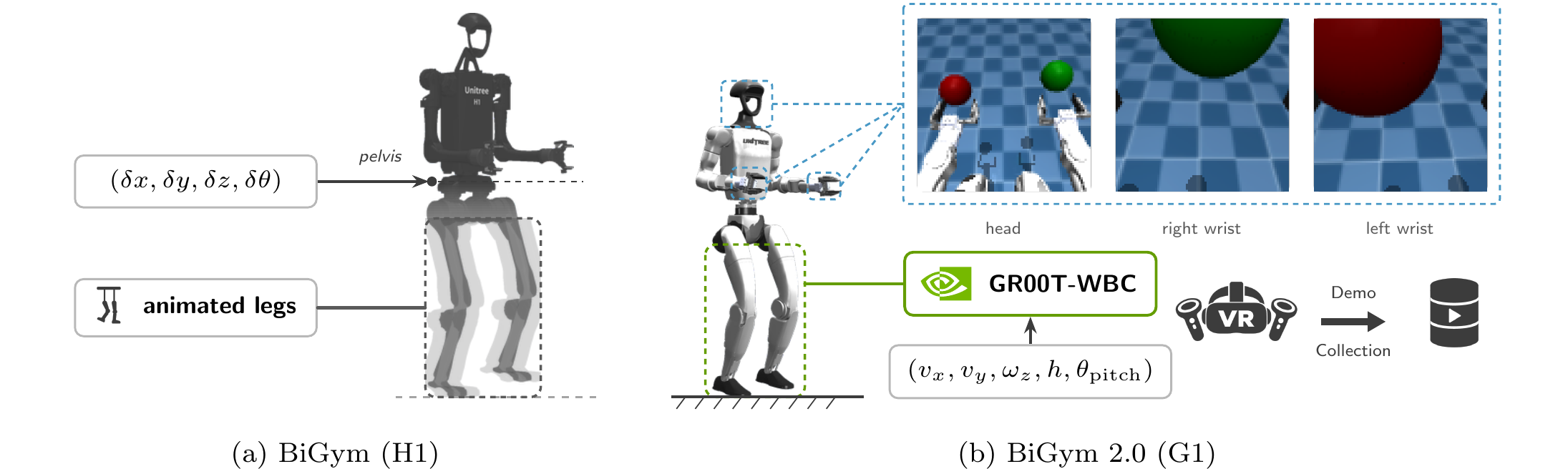}
\else
  \hspace*{0.45cm}\begin{minipage}[t]{5.90cm}\centering\input{figures/fig_diff}\\[3pt]%
    \footnotesize (a) BiGym (H1)\end{minipage}\hfill
  \begin{minipage}[t]{8.85cm}\centering\input{figures/fig_demo}\\[3pt]%
    \footnotesize (b) \benchname{} (G1)\end{minipage}\hspace*{0.45cm}
\fi
  \caption{\textbf{From BiGym to \benchname{}.} (a) BiGym offers full-joint control, but its released demonstrations use the mobile-bimanual mode shown: four increments set the H1 pelvis pose while the legs follow height-conditioned kinematic trajectories. (b) \benchname{} interprets the body commands as velocity, height and, on 14 tasks, torso-pitch references. Frozen GR00T-WBC policies~\cite{nvidiawbc} map them to leg and waist joint targets at 50\,Hz, with the feet supporting the robot. The same controller governs VR demonstration collection, evaluation and reset. Dashed boxes mark three onboard cameras and their recorded views from one dual-reaching episode at the policy input resolution, $84\times84$. The panels are not to relative scale. Arm and gripper commands are omitted.}
  \label{fig:diff}
\end{figure*}
\section{Introduction}
Humanoid robots are expected to perform household tasks that combine walking with manipulation. Carrying an object, reaching into a dishwasher and changing working height require the arms to act while the legs support and move the body. Today, robot capabilities are produced along two distinct routes: data-driven policy learning (VLAs, imitation learning and demo-driven RL) and agentic policy development (coding agents synthesising executable programs from environment APIs). Recent simulation platforms support G1 loco-manipulation through learned controllers~\cite{simple2026,humanoidmimicgen2026,humanoidarena2026}, but no current benchmark combines household tasks, native human demonstrations collected through active whole-body control, deterministic restoration of controller state, and common evaluation of learned and agent-developed policies. This combination matters because both routes must confront the same physical execution stack: unlike a wheeled base that executes velocity commands directly, a walking controller realises them through dynamic gaits that lag, oscillate and perturb the torso~\cite{gysin2003grip,sato2021iros,steadytray2026}. We ask what each route can reliably complete under this execution condition, where their capabilities diverge and where they fail together.

BiGym~\cite{chernyadev2024bigym} provides the closest foundation: 40 H1 household tasks with human virtual reality (VR) demonstrations. It offers a 23-dimensional full-joint whole-body mode and a mobile-bimanual mode, but its released demonstrations use the latter. Its \emph{floating base} directly commands body translation, height and yaw, with planar motion equivalent to an omnidirectional autonomous mobile robot (AMR) base (Fig.~\ref{fig:diff}). This separation isolates mobile-bimanual learning from locomotion control. We instead study legged execution, where gait and balance dynamics couple directly into manipulation. We modify BiGym for the G1 by adapting task geometry, collecting demonstrations through active whole-body control, restoring controller state at evaluation, and supporting learned and agent-developed policies.

Coding agents synthesise robot policies by composing perception, kinematics and control APIs into executable programs~\cite{liang2023code,fu2026capx,elmaaroufi2026rho,lu2026aspire}. We formulate a cold-start capability audit: given environment APIs and three synchronised camera views of a single human demonstration, can a coding agent develop reliable household programs within a bounded interaction budget? To match the learned policies, the agent takes three onboard $84\times84$ views and proprioception as input and outputs the same arm, gripper and body commands, with no camera calibration or inverse-kinematics tools. This cold-start setting measures how reliably an agent develops a program without a pre-built skill library, and gives a baseline against which later work can measure skill accumulation.

We present \benchname{}, a G1 household loco-manipulation benchmark based on BiGym. Our contributions are: (i) adapting \ntasks{} tasks to the G1 workspace, integrating frozen GR00T-WBC walking and balance policies~\cite{nvidiawbc}, and collecting \ndemos{} native human VR demonstrations per task through the same controller used for evaluation (Fig.~\ref{fig:tasks}, Section~\ref{sec:benchmark}); (ii) comparing VLA fine-tuning, imitation learning, online and offline demo-driven RL on nine representative tasks with deterministic evaluation that restores controller states (Table~\ref{tab:main}, Section~\ref{sec:experiments}); (iii) auditing coding agents (GPT-6 Astra and Claude Opus 5.5) under matched observation and action interfaces ($84\times84$ views, direct joint control, no IK tools), showing that their programs lead on bimanual reaching, where a program can control each arm separately, and evaluating a tool-augmented cold-start reference to quantify the benefit of classical robotics tooling (Section~\ref{sec:agent}); and (iv) showing that vision-language-action fine-tuning leads plate and cup transport but not box transport, that multi-object transport remains difficult for imitation learning, demo-driven reinforcement learning and code synthesis, and that cross-workspace stacking remains an open frontier (Sections~\ref{sec:experiments} and~\ref{sec:agent}).

\section{Related Work}
\textbf{Manipulation benchmarks.}
RLBench~\cite{james2020rlbench}, LIBERO~\cite{liu2023libero} and robomimic~\cite{mandlekar2021robomimic} support studies of task variation, learning from demonstrations and transfer, while RoboCasa~\cite{nasiriany2024robocasa}, BEHAVIOR-1K~\cite{li2023behavior1k} and ManiSkill3~\cite{tao2024maniskill3} scale household scenes, activities and simulation throughput. MimicGen~\cite{mandlekar2023mimicgen} and DexMimicGen~\cite{jiang2025dexmimicgen} generate additional demonstrations from a small human set. BiGym~\cite{chernyadev2024bigym} provides 40 household tasks with human demonstrations on the H1. \benchname{} modifies its task layouts and control interface for legged execution on the G1, with demonstrations collected on the adapted system.

\textbf{Whole-body control and teleoperation.}
ExBody~\cite{cheng2024exbody}, HOMIE~\cite{ben2025homie}, AMO~\cite{li2025amo}, TWIST~\cite{ze2025twist} and GR00T~\cite{nvidiawbc} provide interfaces for humanoid whole-body motion. HumanPlus~\cite{fu2024humanplus}, OmniH2O~\cite{he2024omnih2o} and Open-TeleVision~\cite{cheng2024television} support human teleoperation. We use a frozen GR00T-WBC controller during both VR collection and policy evaluation.

\textbf{Humanoid learning environments.}
HumanoidBench~\cite{sferrazza2024humanoidbench} studies locomotion and manipulation with RL, while EgoVLA~\cite{yang2025egovla} provides a humanoid manipulation benchmark with teleoperated data. HumanoidMimicGen~\cite{humanoidmimicgen2026} studies data generation for a nine-task G1 loco-manipulation benchmark. SIMPLE~\cite{simple2026} combines collection and rendering with evaluations of imitation, VLA and world-action model (WAM) policies, and HumanoidArena~\cite{humanoidarena2026} evaluates hierarchical policies under visual, semantic and execution changes. FetchMan~\cite{fetchman2026} studies visual reach-and-pick learning from synthetic demonstrations followed by RL. Humanoid Everyday~\cite{humanoideveryday2025} collects real-robot humanoid manipulation data across everyday activities. Our focus is a suite of household contact and transport tasks with native human demonstrations, supporting comparisons of VLA fine-tuning, imitation learning, demo-driven RL and agent-developed programs. Table~\ref{tab:comparison} summarises the related settings.

\begin{table*}[t]
  \centering\small
  \caption{\textbf{Humanoid manipulation benchmarks and demonstration resources.} Task counts and reported policy families follow the cited papers. The two demonstration columns distinguish the availability of demonstrations from that of human demonstrations. Low-level execution names the interface or controller, not whether it is updated with the task policy.}
  \label{tab:comparison}
  \setlength{\tabcolsep}{6pt}
\sbox{\tablebox}{%
\begin{tabular}{@{}llclccl@{}}
\toprule
 & & & & \multicolumn{2}{c}{Demonstrations} & \\
\cmidrule(lr){5-6}
Benchmark & Robot & Tasks & Low-level execution & Any & Human & Policy families \\
\midrule
HumanoidBench~\cite{sferrazza2024humanoidbench} & H1 / Digit & 27\rlap{$^a$} & Full joints & -- & -- & RL \\
BiGym~\cite{chernyadev2024bigym} & H1 & 40 & FB / full joints\rlap{$^b$} & $\checkmark$ & $\checkmark$ & IL, RL \\
Ego Humanoid Manipulation~\cite{yang2025egovla} & H1 & 12 & Arm IK + hand PD & $\checkmark$ & $\checkmark$ & IL, VLA \\
HumanoidMimicGen~\cite{humanoidmimicgen2026} & G1 & 9 & Learned controller & $\checkmark$ & $\checkmark$\rlap{$^c$} & IL, VLA \\
SIMPLE~\cite{simple2026} & G1 & 60\rlap{$^d$} & AMO / SONIC WBC & $\checkmark$ & $\checkmark$\rlap{$^c$} & IL, VLA, WAM \\
HumanoidArena~\cite{humanoidarena2026} & G1 & 7 & Learned trackers & $\checkmark$ & $\checkmark$ & IL, VLA \\
\midrule
\textbf{\benchname{} (ours)} & G1 & \ntasks{} & GR00T-WBC & $\checkmark$ & $\checkmark$ & IL, RL, VLA, agent \\
\bottomrule
\end{tabular}}%
\usebox{\tablebox}
\par\smallskip
\begin{minipage}{\wd\tablebox}
\footnotesize
\zhen{
$^a$12 个行走任务、15 个操作任务。
$^b$FB：浮动底座；BiGym 的移动双臂实验与示教采用此配置。
$^c$同时包含人工与生成示教。
$^d$主对比评测 6 个任务，另有 6 个任务使用单一策略评测。
}{
$^a$12 locomotion and 15 manipulation tasks.
$^b$FB: floating base, used for BiGym's mobile-bimanual experiments and demonstrations.
$^c$Includes both human and generated demonstrations.
$^d$Six tasks in the main comparison; six further tasks evaluated with one policy.
}
\end{minipage}

\end{table*}

\textbf{Policy learning and program development.}
Pretrained VLAs such as OpenVLA~\cite{kim2024openvla}, \pizero{}~\cite{pi05} and GR00T N1~\cite{nvidia2025groot} can be fine-tuned on demonstrations, while ACT~\cite{zhao2023act} and DP~\cite{chi2023diffusion} learn action sequences by imitation learning. CQN-AS~\cite{seo2025cqnas}, which extends coarse-to-fine RL~\cite{seo2024cqn} to action sequences, and \drq{}~\cite{yarats2022drqv2,seo2024cqn} combine demonstrations with online interaction. DEAS~\cite{deas2025} learns from a fixed offline dataset.

Language models generating robot policy code have an established research basis~\cite{liang2023code}. CaP-X studies how interface abstraction, interaction and perception affect code-policy performance~\cite{fu2026capx}. RHO uses reflection and search to optimise policy repositories during development, deploying frozen code in its CaP-Bench experiments~\cite{elmaaroufi2026rho}. ASPIRE uses fine-grained execution traces to diagnose failures, validate repairs and accumulate reusable skills~\cite{lu2026aspire}. We follow this code-policy route to benchmark programs developed by GPT-6 Astra and Claude Opus 5.5 within a charged interaction budget on G1 household loco-manipulation. We isolate cold-start policy synthesis per task to measure development reliability, while using the suite to define the progression toward cumulative skill reuse. Section~\ref{sec:policies} specifies the agent protocol and Section~\ref{sec:agent} the tool interfaces.

\section{\benchname}
\label{sec:benchmark}
\subsection{Loco-manipulation challenges and tasks}
We organise \benchname{} around two physical challenges and one evaluation question:
(i) \emph{Gait-perturbed interaction}: arms act while legs step and balance, coupling manipulation to dynamic footfall shock and base lag;
(ii) \emph{Coordinated multi-object transport}: carrying items across tables and counters demands continuous spatial tracking across changing visual viewpoints;
(iii) \emph{Cross-paradigm reliability}: how reliably learned policies and agent-written programs complete tasks when they share the same observations, actions and controller.

The suite instantiates these challenges across \ntasks{} tasks in four scene families (Fig.~\ref{fig:tasks}):

\textbf{Reaching (3 tasks).} Single reaching specifies the left hand, multi-modal reaching permits either hand, and dual reaching requires both hands to reach targets simultaneously, separating hand selection from bimanual coordination.

\textbf{Tabletop (5 tasks).} Single- and two-plate transport move plates between drying racks. Cup and cutlery flipping require reorientation. Block stacking combines cross-workspace transport with successive precise placements between tables.

\textbf{Dishwasher (4 tasks).} Door closing requires pushing racks and closing the door. Loading tasks place cups in the upper tray, cutlery in the basket and plates in the lower rack, combining fixture handling with constrained placement.

\textbf{Kitchen counter (8 tasks).} Four drawer and cupboard tasks provide fixture manipulation. Box pickup transports a large parcel to the counter. Further tasks store cups in cabinets, move a saucepan to the hob and transfer a sandwich with a spatula.

We adapt BiGym's contact geometry to the G1 workspace. Plate racks are lowered and moved inward, and the dishwasher door/tray scene uses a 0.25\,m plinth. For box pickup, the side table and counter are 0.50\,m and 0.71\,m high, respectively. These adapted layouts define the benchmark tasks.

Success is defined by the achieved scene state. Plate placement requires release, contact with the destination rack and the specified pose tolerance. Closing the dishwasher requires both trays and the door to be closed. Stacking requires a released three-block contact chain with height and planar-alignment constraints. These conditions must persist for 1\,s. \supp{Appendix~\ref{app:tasks} gives the full predicates and reset distributions.}{The released benchmark specifies the full predicates and reset distributions.}

\subsection{Whole-body execution}
The robot is a 29-DoF Unitree G1 with two \mbox{Dex1-1} parallel grippers in MuJoCo~\cite{todorov2012mujoco}. A high-level policy sends arm joint targets, gripper commands and body-motion commands. Frozen GR00T-WBC balance and walking policies convert the body commands into lower-body joint targets. The controller maintains balance while executing locomotion, height changes and torso pitch during manipulation.

Body commands specify forward and lateral velocity, yaw velocity and height, with optional torso pitch. Together with fourteen arm targets and two gripper commands, they form a 20- or 21-dimensional action. The high-level interface runs at 50\,Hz, with 1\,ms physics integration in the G1 scene. Reset restores simulator and controller state, then performs 200 high-level warmup steps. A Gymnasium interface supports demo-driven agent training (Fig.~\ref{fig:api}). \supp{Action layouts, bounds and controller settings are specified in Appendix~\ref{app:interfaces}.}{Action layouts, bounds and controller settings are specified with the released code.}

\begin{figure}[t]
\centering
\input{figures/fig_api}
\caption{\textbf{Training an agent with \benchname{}.} Load demonstrations, interact with the environment and update the agent through the Gymnasium API. The whole-body controller executes body commands within each environment step. \texttt{Agent} represents the user\textquotesingle s learning algorithm.}
\label{fig:api}
\end{figure}

\subsection{Human demonstrations}
An operator supplies hand targets through VR and Mink inverse kinematics (IK)~\cite{Zakka_Mink_Python_inverse_2026}, and high-level body commands through thumbsticks. GR00T-WBC executes the body commands and maintains balance throughout collection. Collection requires the success condition to hold for 3\,s. The training view retains the first 1\,s of this hold, matching evaluation.

Each episode stores synchronised RGB, robot proprioception, normalised high-level actions, rewards and terminal indicators. Simulator state and controller snapshots are retained for replay and re-rendering, separately from the visual policy's inputs. A native MuJoCo viewer and a Viser-based web viewer~\cite{yi2025viser} support episode selection and frame stepping.

\begin{figure}[t]
\centering
\includegraphics[width=\columnwidth]{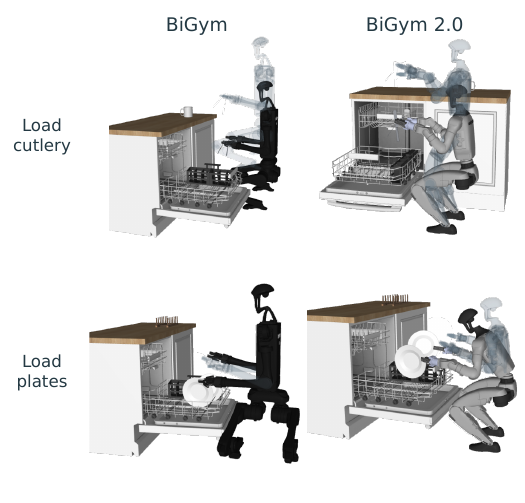}
\caption{\textbf{Body posture during cutlery and plate loading.} Left: BiGym; right: \benchname{}. Each panel overlays two states from one demonstration, with the earlier pose faint and the later pose opaque. Both systems adjust working height. G1 also pitches its torso forward during plate loading. Robots and task layouts follow their respective benchmarks.}
\label{fig:demoposes}
\end{figure}

The suite provides \ndemos{} successful demonstrations for each of the \ntasks{} tasks, 1,200 episodes in total. As a loco-manipulation benchmark, \benchname{} spans near-stationary manipulation and cross-workspace transport, with task-wise median planar base travel ranging from 0.06 to 8.10\,m (5\,Hz, 5\,cm dead band). Tasks also involve changes in working height and torso posture. Fig.~\ref{fig:demoposes} illustrates height adjustment during cutlery loading and G1 torso pitch during plate loading. \supp{Appendix~\ref{app:data} gives the trajectory statistics and recording format.}{The release documents the trajectory statistics and recording format.}

\subsection{Observations and evaluation}
In the main comparison, every method observes the same three $84\times84$ head/wrist RGB views and the robot's proprioception, and none receives object poses or other simulator state. ACT, DP, CQN-AS, DrQ-v2+ and DEAS stack four frames. \pizero{} additionally receives a task instruction. Under the strict interface, coding-agent programs observe the views and the 50- or 56-dimensional proprioceptive vector at run time, and send body, arm and gripper commands to the same controller in physical units\supp{ (bounds in Appendix~\ref{app:agent})}{}. They receive no camera calibration, ray projection, inverse kinematics, privileged wrist positions or rate limiters. Section~\ref{sec:agent} separately evaluates a tool-augmented interface that adds these tools.

Each non-VLA learned-policy checkpoint is evaluated on \neps{} episodes with seeds $\seedbase+i$. Dishwasher closing and the two wall-cupboard tasks reset to a fixed initial state and do not consume these seeds. The other tasks randomise object placement per seed. Reset determinism is measured rather than assumed: without controller-state restoration, a seeded reset of a used environment differs from a fresh one by $9.5\times10^{-2}$ in configuration and by $2.1\times10^{-1}$ after 50 steps. With it enabled, the trajectories are bit-identical on the released substrate build. In the main comparison, the reported score averages the last five evaluated checkpoints of each run, then reports the mean and standard error across \nseeds{} independent runs~\cite{henderson2018deeprl,agarwal2021rliable,colas2018seeds,snyder2025imitation}. The nominal window is 80k--100k. \supp{Selecting each run's best checkpoint would instead consult the evaluation seeds (Appendix~\ref{app:analysis}).}{Learning curves and per-run results are released with the benchmark.} The \pizero{} scores pool the last five checkpoints (26k--30k) with 50 episodes each, from one training run per task.

\subsection{Policy families}
\label{sec:policies}
For VLA fine-tuning, \pizero{}~\cite{pi05} uses supervised fine-tuning on the converted demonstrations and task instructions, with relative-delta actions and 16 executed actions per prediction. ACT~\cite{zhao2023act} and Diffusion Policy (DP)~\cite{chi2023diffusion} learn action sequences by imitation learning. Both train for 101k updates with batch size 256.

CQN-AS~\cite{seo2025cqnas}, \drq{}~\cite{yarats2022drqv2,seo2024cqn} and DEAS~\cite{deas2025} are demo-driven RL. The first two learn online with demonstrations, using 101k environment interactions and batches of 256 replay and 256 demonstration transitions. Successful online episodes can enter the demonstration buffer. DEAS learns offline over action sequences, using 101k updates on the fixed dataset. Our pixel-based implementation uses advantage-weighted regression (AWR) policy extraction. The online and offline budgets are counted in different units---interactions against updates---and are reported as such.

The reference execution recipe replans after each action for ACT, after eight actions for DP and after sixteen for DEAS. At 50\,Hz these intervals are 0.02, 0.16 and 0.32\,s. The non-VLA recipe uses random image shifts with padding 8 and masks non-base proprioception with probability 0.2 during training. \supp{Full method configurations are given in Appendix~\ref{app:baselines}.}{Full method configurations ship with the released code.}

\begin{figure*}[t]
\centering
\iflatexml
\includegraphics[width=\textwidth]{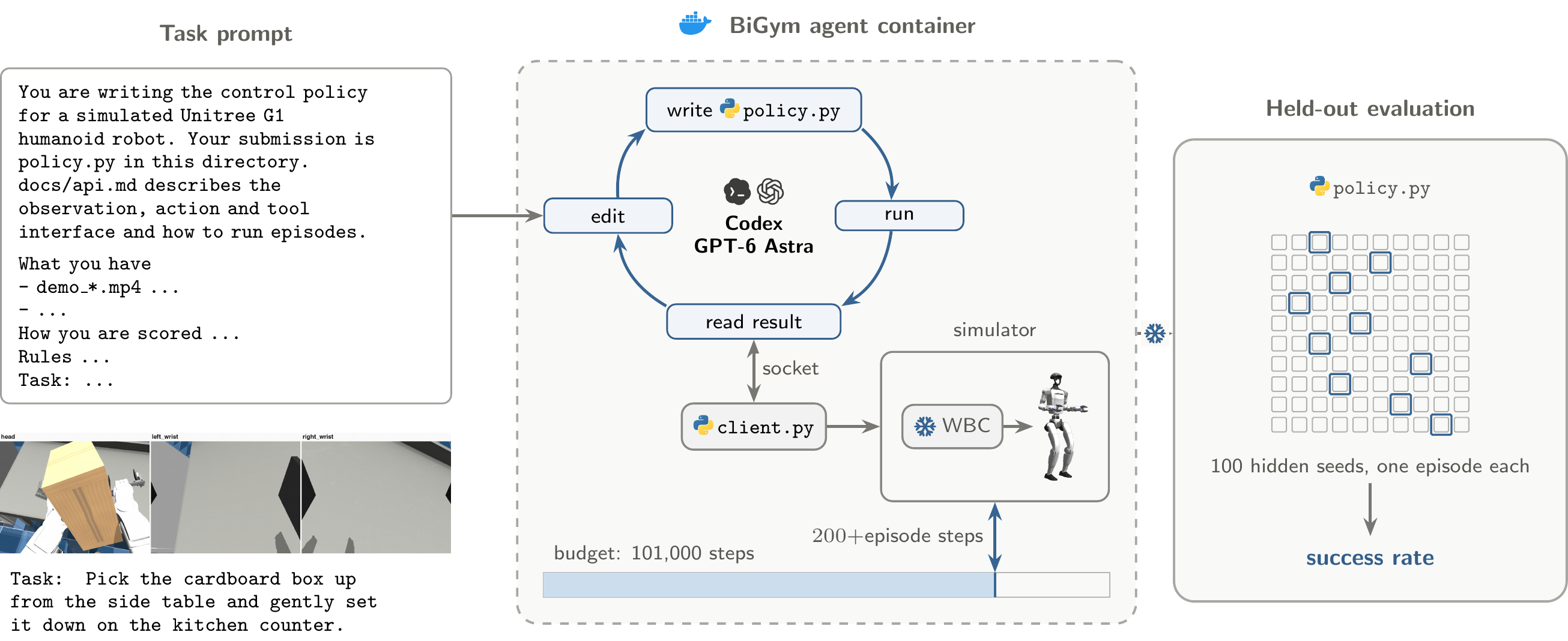}
\else
\resizebox{\textwidth}{!}{\input{figures/fig4_agent}}
\fi
\caption{\textbf{Coding-agent development from a single demonstration.} The agent receives three synchronised camera views of one human demonstration and onboard camera access, developing a program under the strict interface within a shared interaction budget. The frozen program is evaluated on 100 hidden seeds, with ten episodes repeated to check deterministic outcomes.}
\label{fig:agent-main}
\end{figure*}

GPT-6 Astra and Claude Opus 5.5 develop Python policies from a cold start (Fig.~\ref{fig:agent-main}). For each task, a fresh session receives a task description, the environment API and three synchronised camera views of one demonstration, and carries nothing over from other tasks. A session may use 101,000 environment steps, the same count as the online learners, with 200 charged per reset. Development uses only the onboard views and the \ndemos{} demonstration seeds. The prompt allows NumPy, SciPy, OpenCV and Pillow, and forbids training networks, fitting regressions or building demonstration lookup tables. The submitted program runs without LLM calls on seeds $\seedbase+i$, which the agent never sees, and ten episodes are repeated to check that outcomes are deterministic. Each agent has three sessions per task.

\section{Capability audit}
\label{sec:experiments}
Table~\ref{tab:main} audits learned policies and coding-agent programs across nine representative tasks under identical whole-body execution. Under the strict interface, \pizero{} has the highest nine-task mean, above imitation learning and both coding agents. Both coding agents still beat every demo-driven RL mean and rank alongside imitation learning. Opus leads multi-modal reaching. On dual reaching, both coding agents stay above every learned policy. \pizero{} leads single-plate transport, two-plate transport and cup loading. Pick-box transport remains higher for Opus and ACT than for \pizero{}. Cross-workspace stacking was not evaluated for \pizero{} and remains unsolved by ACT and CQN-AS.

\begin{table*}[t]
  \centering\footnotesize
  \caption{\textbf{Success rates (\%) on nine representative tasks.} ACT, DP, CQN-AS, DrQ-v2+ and DEAS: mean $\pm$ standard error (SE) across three training runs, each averaging its last five checkpoints with 100 episodes each. \pizero{}: one training run per task, pooled over its last five checkpoints with 50 episodes each. Coding agents: three independent development sessions each of Claude Opus 5.5 and GPT-6 Astra under the strict interface ($84\times84$ onboard views, direct joint actions, no kinematics tools), 100 episodes per frozen program.}
  \label{tab:main}
  {\setlength{\tabcolsep}{3pt}\begin{tabular}{@{}lcccccccc@{}}
\toprule
 & VLA & \multicolumn{2}{c}{Imitation learning} & \multicolumn{3}{c}{Demo-driven RL} & \multicolumn{2}{c}{Coding agent (strict)} \\
\cmidrule(lr){2-2} \cmidrule(lr){3-4} \cmidrule(lr){5-7} \cmidrule(lr){8-9}
Task & $\pi_{0.5}$ & ACT & DP & CQN-AS & DrQ-v2+ & DEAS & Opus 5.5 & Astra \\
\midrule
Pick box & \phantom{0}19 & \phantom{0}61 $\pm$ 4 & \phantom{0}10 $\pm$ 3 & \phantom{0}48 $\pm$ 2 & \phantom{00}0 $\pm$ 0 & \phantom{0}14 $\pm$ 3 & \textbf{\phantom{0}71 $\pm$ 16} & \phantom{0}54 $\pm$ 22 \\
Reach multi-modal & \phantom{0}90 & \phantom{0}81 $\pm$ 1 & \phantom{0}84 $\pm$ 1 & \phantom{0}19 $\pm$ 16 & \phantom{00}0 $\pm$ 0 & \phantom{0}79 $\pm$ 2 & \textbf{100 $\pm$ 0} & \phantom{0}89 $\pm$ 7 \\
Reach dual & \phantom{0}58 & \phantom{0}30 $\pm$ 0 & \phantom{0}27 $\pm$ 0 & \phantom{00}3 $\pm$ 1 & \phantom{00}0 $\pm$ 0 & \phantom{0}34 $\pm$ 1 & \phantom{0}78 $\pm$ 10 & \textbf{\phantom{0}89 $\pm$ 7} \\
Drawer close & \textbf{100} & \textbf{100 $\pm$ 0} & \textbf{100 $\pm$ 0} & \textbf{100 $\pm$ 0} & \phantom{0}99 $\pm$ 0 & \textbf{100 $\pm$ 0} & \textbf{100 $\pm$ 0} & \phantom{0}99 $\pm$ 1 \\
Drawer open & \textbf{\phantom{0}99} & \phantom{0}92 $\pm$ 0 & \phantom{0}98 $\pm$ 0 & \phantom{0}89 $\pm$ 2 & \phantom{00}0 $\pm$ 0 & \phantom{0}79 $\pm$ 4 & \textbf{\phantom{0}99 $\pm$ 1} & \textbf{\phantom{0}99 $\pm$ 1} \\
Move plate & \textbf{\phantom{0}65} & \phantom{0}60 $\pm$ 1 & \phantom{0}62 $\pm$ 4 & \phantom{0}48 $\pm$ 3 & \phantom{00}0 $\pm$ 0 & \phantom{0}27 $\pm$ 3 & \phantom{0}20 $\pm$ 2 & \phantom{0}14 $\pm$ 11 \\
Move two plates & \textbf{\phantom{0}68} & \phantom{0}39 $\pm$ 0 & \phantom{0}18 $\pm$ 4 & \phantom{0}21 $\pm$ 3 & \phantom{00}0 $\pm$ 0 & \phantom{00}2 $\pm$ 1 & \phantom{0}15 $\pm$ 4 & \phantom{0}16 $\pm$ 11 \\
Dishwasher close & \phantom{0}79 & \phantom{0}93 $\pm$ 5 & \phantom{0}51 $\pm$ 2 & \phantom{0}78 $\pm$ 1 & \phantom{00}0 $\pm$ 0 & \phantom{0}27 $\pm$ 3 & \phantom{0}33 $\pm$ 33 & \textbf{100 $\pm$ 0} \\
Dishwasher load cups & \textbf{\phantom{0}92} & \phantom{0}56 $\pm$ 3 & \phantom{0}51 $\pm$ 0 & \phantom{0}47 $\pm$ 7 & \phantom{00}0 $\pm$ 0 & \phantom{00}6 $\pm$ 2 & \phantom{0}28 $\pm$ 27 & \phantom{0}21 $\pm$ 4 \\
\midrule
Mean & \textbf{\phantom{0}75} & \phantom{0}68 & \phantom{0}56 & \phantom{0}50 & \phantom{0}11 & \phantom{0}41 & \phantom{0}61 & \phantom{0}65 \\
\bottomrule
\end{tabular}
}
\end{table*}

\subsection{Dual reaching and transport separate the methods}
Every method solves drawer closing. Dual reaching separates the paradigms: every non-zero learned policy drops sharply from multi-modal reaching, where either hand suffices, to both hands at once. Both coding agents stay above every learned policy on dual reaching. On multi-modal reaching, Opus remains ahead of every learned policy, while Astra falls just below \pizero{}. All six agent programs for dual reaching control the two arms separately, each with its own target and its own correction. A learned policy must instead reproduce the bimanual coordination from demonstrations.

Adding a second plate lowers success for ACT, DP, CQN-AS and DEAS, but not for \pizero{}. Coding agents succeed on few plate-transport or cup-loading episodes. Opus succeeds on most pick-box episodes. No learned method leads on all nine tasks.\supp{ Appendix~\ref{app:qualitative} describes the contact fix for gait-induced grip slip and a \drq{} failure mode.}{}

\subsection{Checkpoint selection changes the ranking}
On single-plate transport, among the three-run methods, ACT has the highest mean peak checkpoint while DP has the highest last-five mean, and \pizero{}'s last-five score is higher than that DP mean. Selecting the best of 20 checkpoints, each estimated from 100 episodes, takes a maximum over noisy estimates. \supp{Table~\ref{tab:appendix-peak-final} lists peak and final scores per method.}{Per-run peak and final scores are released with the benchmark.}

\subsection{The wider suite presents further challenges}
Table~\ref{tab:coverage} checks that the remaining eleven tasks are learnable from the provided demonstrations: ACT or CQN-AS reaches at least 10\% on every task except cross-workspace stacking, which neither method solves in more than 1\% of episodes. Section~\ref{sec:agent} examines the program policies and development budget for stacking.

\begin{table}[t]
\centering\small
\caption{\textbf{ACT and CQN-AS success rates (\%) on the remaining eleven tasks.} Mean $\pm$ SE across three training runs, each averaging its last five checkpoints with 100 episodes each.}
\label{tab:coverage}
\begin{tabular}{@{}lcc@{}}
\toprule
Task & ACT & CQN-AS \\
\midrule
Reach single & \textbf{\phantom{0}89 $\pm$ 0} & \phantom{0}84 $\pm$ 1 \\
Flip cup & \textbf{\phantom{0}62 $\pm$ 1} & \phantom{0}42 $\pm$ 9 \\
Flip cutlery & \phantom{0}46 $\pm$ 1 & \textbf{\phantom{0}55 $\pm$ 2} \\
Stack blocks & \phantom{00}0 $\pm$ 0 & \textbf{\phantom{00}1 $\pm$ 0} \\
Load cutlery & \textbf{\phantom{0}11 $\pm$ 1} & \phantom{0}10 $\pm$ 3 \\
Load plates & \textbf{\phantom{0}60 $\pm$ 1} & \phantom{0}34 $\pm$ 9 \\
Put cups & \phantom{0}38 $\pm$ 1 & \textbf{\phantom{0}49 $\pm$ 2} \\
Saucepan to hob & \phantom{0}19 $\pm$ 1 & \textbf{\phantom{0}32 $\pm$ 5} \\
Remove sandwich & \textbf{\phantom{0}38 $\pm$ 2} & \phantom{0}17 $\pm$ 2 \\
Wall cupboard close & \textbf{100 $\pm$ 0} & \phantom{0}60 $\pm$ 31 \\
Wall cupboard open & \textbf{\phantom{0}87 $\pm$ 7} & \phantom{00}7 $\pm$ 7 \\
\bottomrule
\end{tabular}

\end{table}

\section{Coding-agent policies}
\label{sec:agent}
Table~\ref{tab:main} reports both agents under the strict interface and the protocol of Section~\ref{sec:policies}. This section examines how their programs behave and what classical robotics tools add.

\subsection{Harnesses and interfaces}
Each model runs in its official harness (Codex CLI v0.153.4, Claude Code v2.1.280) at high reasoning effort, in an isolated container with network access only to the model API\supp{ (Appendix~\ref{app:agent})}{}.

\supp{The strict interface withholds tools, not the prior knowledge a model brings. Claude Opus 5.5 solved IK numerically in 24 of its 27 strict programs, and 22 of them use G1 link offsets recalled from Unitree's public robot description rather than read from the simulator (Appendix~\ref{app:agent-knowledge}). No GPT-6 Astra program does either. The two agents reach similar nine-task means (Table~\ref{tab:main}), so recalled kinematics did not remove the failures on contact-rich tasks.}{}

The tool-augmented interface adds $640\times480$ views, camera calibration, pixel-to-ray projection, differential IK, rate limiters and world-frame wrist and base positions. We evaluate it for Astra only, in a separate set of sessions on a pre-release build that develop on seeds 0--199 (Table~\ref{tab:agent-tools}), to measure what classical robotics tools add to code synthesis.

Over its 27 strict sessions, Opus 5.5 read about four times as many tokens as Astra and cost more (\$420 versus \$296, API-equivalent). Per-session ledgers are released. In the pre-release sessions of Table~\ref{tab:agent-tools}, Astra's strict sessions cost more than its tool-augmented ones, consistent with more trial and error without kinematics solvers.

\subsection{Visual servoing under strict constraints, but contact vulnerability}
Under the strict interface, both agents outperform the demo-driven RL methods on the nine-task mean. Astra's programs show two execution modes: on reaching tasks, they close the loop in pixel space, thresholding the target in the wrist camera and moving the shoulder joints in proportion to its pixel offset, without 3D coordinates. For drawer closing, they replay an open-loop joint sequence that shoves the drawer shut without visual feedback, exploiting fixed contact geometry.

Both agents fail most plate-transport and cup-loading episodes, and results vary widely across sessions: a task is often solved in one session and failed in the others\supp{ (Appendix~\ref{app:agent-knowledge})}{}, so a three-session mean reflects how often development finds a working strategy.

\subsection{Impact of robotics tooling and cross-workspace stacking}
Table~\ref{tab:agent-tools} compares the two interfaces. The full toolchain raises the nine-task mean, but unevenly: reaching and drawer closing barely change, so pixel-space servoing already suffices for open-space alignment, whereas drawer opening, dishwasher closing and cup loading gain the most.

Block stacking requires carrying blocks 0.98\,m from one table to a pad on another and making successive precise placements. All 60 demonstrations contain two outbound carries and one return, with a median planar base path of 8.1\,m. Each block must rest on the one below within a 5\,cm planar tolerance. The released stack must hold for one second, and any block touching the floor terminates the episode. Under the tool-augmented interface, three Astra sessions reach 19\%, 0\% and 0\%. The strongest program stores world-frame block and pad coordinates during initial perception and navigates closed-loop on the proprioceptive base pose, while grasp transitions remain time-indexed. With the 101,000-step budget and an 8,750-step episode limit, trials running to completion admit only eleven episodes, leaving narrow margins to repair long-horizon programs under bounded interaction.

\begin{table}[t]
\centering\small
\caption{\textbf{Coding-agent (GPT-6 Astra) performance and development cost.} Strict vs.\ tool-augmented interfaces across nine tasks, from a separate set of sessions on a pre-release build of the environment, so strict scores differ from Table~\ref{tab:main}. Task costs report mean per-session expenditure, with 27-session totals below.}
\label{tab:agent-tools}
{\setlength{\tabcolsep}{2.5pt}\begin{tabular}{@{}l r@{\,$\pm$\,}l r@{\,$\pm$\,}l r rr@{}}
\toprule
 & \multicolumn{5}{c}{Success rate (\%)} & \multicolumn{2}{c}{Cost/session (\$)} \\
\cmidrule(lr){2-6} \cmidrule(lr){7-8}
Task & \multicolumn{2}{c}{Strict} & \multicolumn{2}{c}{Tools} & $\Delta$ & Strict & Tools \\
\midrule
Reach multi      & 100 & 0  & 100 & 0  &  0 & \phantom{0}3.2 & \phantom{0}2.1 \\
Reach dual       & \phantom{0}94 & 2  & \phantom{0}99 & 1  &  5 & \phantom{0}5.3 & \phantom{0}5.2 \\
Drawer close     & 100 & 0  & 100 & 0  &  0 & \phantom{0}1.8 & \phantom{0}2.3 \\
Drawer open      & \phantom{0}55 & 27 & \phantom{0}99 & 1  & 44 &           10.5 & \phantom{0}5.2 \\
Move plate       & \phantom{0}39 & 5  & \phantom{0}41 & 26 &  2 &           10.8 &           11.7 \\
Move 2 plates    & \phantom{00}0$^*$ & 0  & \phantom{0}27 & 25 & 27 &           25.3 &           15.4 \\
Pick box         & \phantom{0}24 & 20 & \phantom{0}34 & 13 & 11 &           18.2 & \phantom{0}9.9 \\
Dishwasher close & \phantom{0}33 & 33 & 100 & 0  & 67 &           19.7 &           16.2 \\
Load cups        & \phantom{0}32 & 16 & \phantom{0}85 & 8  & 53 &           14.0 &           10.4 \\
\midrule
Mean             & \multicolumn{2}{c}{\textbf{53}} & \multicolumn{2}{c}{\textbf{76}} & \textbf{23} & \textbf{12.1} & \textbf{\phantom{0}8.7} \\
Total (27 sessions)  & \multicolumn{5}{c}{---} & \textbf{326.3} & \textbf{235.1} \\
\midrule[\heavyrulewidth]
\multicolumn{8}{@{}l}{\scriptsize $^*$Reflects 1/300 successful episodes (0.33\%, rounded to 0\%).} \\
\bottomrule
\end{tabular}
}
\end{table}

\section{Discussion and limitations}
\label{sec:limitations}
Both agent interfaces are cold-start: each task is developed in a fresh session that carries over no code, skills or notes from other tasks. We chose this to match the learned policies, each trained on a single task, so that neither route draws on other \benchname{} tasks. Within a task, an agent sees one demonstration, whereas a learned policy trains on \ndemos{}. We did not test whether agents improve by accumulating verified programs and reusing them across tasks. Recent work on fixed-base manipulation shows that such accumulation improves cross-task generalisation and reduces synthesis cost~\cite{lu2026aspire}. Our agent results are therefore a cold-start reference rather than the ceiling of agent development, and the large variance across sessions\supp{ (Appendix~\ref{app:agent-knowledge})}{} is one cost that accumulation may reduce. \benchname{} provides the testbed to measure whether these gains carry over to humanoid loco-manipulation under whole-body control.

All evaluations use simulation with one frozen whole-body controller per robot, characterising the joint policy--controller stack. Dishwasher closing and wall-cupboard tasks reset to fixed initial states as deterministic execution checks, while other tasks randomise object placements per seed. Because policies command high-level base velocities and pelvis height references rather than joint torques, execution inherently couples to the controller tracking accuracy and posture regulation under contact disturbances. Evaluating alternative whole-body controllers would help assess how policy performance depends on controller-specific dynamics. \supp{Agent results also rest on pretraining priors: Claude Opus 5.5 recalled the published kinematics of the G1, a well-documented robot (Section~\ref{sec:agent}), so agents may do worse on embodiments for which they hold no such prior. }{}Future work includes diverse walking controllers, world-action models~\cite{ye2026dreamzero} and physical deployment.

\section{Conclusion}
We present \benchname{}, a 20-task household loco-manipulation benchmark adapting BiGym to the Unitree G1 with native human demonstrations under closed-loop whole-body control. We compare vision-language-action fine-tuning (\pizero{}), imitation learning, demo-driven reinforcement learning, and cold-start coding agents (GPT-6 Astra and Claude Opus 5.5) under the same execution constraints. Vision-language-action fine-tuning attains the highest nine-task mean and leads plate and cup transport; agent-developed programs lead on bimanual reaching, where a program can control each arm separately. Cross-workspace stacking remains open, \pizero{} stays low on pick-box, and multi-object transport remains difficult for imitation learning, demo-driven reinforcement learning and agent-developed policies. Evaluation restores controller state, so results are reproducible, and the cold-start agent results give a baseline for work on skill accumulation. We plan to maintain and extend the suite.

\FloatBarrier
\phantomsection\label{paper:body-end}
\bibliographystyle{IEEEtran}
\bibliography{IEEEabrv,bib/refs}

\begin{thebibliography}{10}
\providecommand{\url}[1]{#1}
\csname url@samestyle\endcsname
\providecommand{\newblock}{\relax}
\providecommand{\bibinfo}[2]{#2}
\providecommand{\BIBentrySTDinterwordspacing}{\spaceskip=0pt\relax}
\providecommand{\BIBentryALTinterwordstretchfactor}{4}
\providecommand{\BIBentryALTinterwordspacing}{\spaceskip=\fontdimen2\font plus
\BIBentryALTinterwordstretchfactor\fontdimen3\font minus
  \fontdimen4\font\relax}
\providecommand{\BIBforeignlanguage}[2]{{%
\expandafter\ifx\csname l@#1\endcsname\relax
\typeout{** WARNING: IEEEtran.bst: No hyphenation pattern has been}%
\typeout{** loaded for the language `#1'. Using the pattern for}%
\typeout{** the default language instead.}%
\else
\language=\csname l@#1\endcsname
\fi
#2}}
\providecommand{\BIBdecl}{\relax}
\BIBdecl

\bibitem{nvidiawbc}
{NVIDIA}, ``{GR00T Whole-Body Control},''
  \url{https://github.com/NVlabs/GR00T-WholeBodyControl}, accessed: 2026-09-10.

\bibitem{simple2026}
S.~Wei \emph{et~al.}, ``{SIMPLE: Simulation-Based Policy Learning and
  Evaluation for Humanoid Loco-manipulation},'' \emph{arXiv preprint
  arXiv:2606.08278}, 2026.

\bibitem{humanoidmimicgen2026}
K.~Lin \emph{et~al.}, ``{HumanoidMimicGen: Data Generation for
  Loco-Manipulation via Whole-Body Planning},'' \emph{arXiv preprint
  arXiv:2605.27724}, 2026.

\bibitem{humanoidarena2026}
T.~Wang \emph{et~al.}, ``{HumanoidArena: Benchmarking Egocentric Hierarchical
  Whole-body Learning},'' \emph{arXiv preprint arXiv:2606.17833}, 2026.

\bibitem{gysin2003grip}
P.~Gysin, T.~R. Kaminski, and A.~M. Gordon, ``{Coordination of fingertip forces
  in object transport during locomotion},'' \emph{Exp. Brain Res.}, vol. 149,
  no.~3, pp. 371--379, 2003.

\bibitem{sato2021iros}
S.~Sato \emph{et~al.}, ``{Drop Prevention Control for Humanoid Robots Carrying
  Stacked Boxes},'' in \emph{Proc. IEEE/RSJ Int. Conf. Intell. Robots Syst.
  (IROS)}, 2021, pp. 4118--4125.

\bibitem{steadytray2026}
A.~Huang, Z.~Wu, S.~Atar, Y.~Zhi, and M.~Yip, ``{SteadyTray: Learning Object
  Balancing Tasks in Humanoid Tray Transport via Residual Reinforcement
  Learning},'' \emph{arXiv preprint arXiv:2603.10306}, 2026.

\bibitem{chernyadev2024bigym}
N.~Chernyadev, N.~Backshall, X.~Ma, Y.~Lu, Y.~Seo, and S.~James, ``{BiGym: A
  Demo-Driven Mobile Bi-Manual Manipulation Benchmark},'' in \emph{Proc. Conf.
  Robot Learn. (CoRL)}, ser. PMLR, vol. 270, 2024, pp. 4201--4217.

\bibitem{liang2023code}
J.~Liang \emph{et~al.}, ``Code as policies: Language model programs for
  embodied control,'' in \emph{Proc. IEEE Int. Conf. Robot. Autom. (ICRA)},
  2023.

\bibitem{fu2026capx}
L.~Fu \emph{et~al.}, ``{CaP-X: A Framework for Benchmarking and Improving
  Coding Agents for Robot Manipulation},'' \emph{arXiv preprint
  arXiv:2603.22435}, 2026.

\bibitem{elmaaroufi2026rho}
K.~Elmaaroufi, J.~Svegliato, S.~Kalade, G.~Schelle, S.~A. Seshia, and
  M.~Zaharia, ``{RHO: Your Coding Agent is Secretly a Roboticist},''
  \emph{arXiv preprint arXiv:2606.16458}, 2026.

\bibitem{lu2026aspire}
R.~Lu \emph{et~al.}, ``{ASPIRE: Agentic /Skills Discovery for Robotics},''
  \emph{arXiv preprint arXiv:2607.00272}, 2026.

\bibitem{james2020rlbench}
S.~James, Z.~Ma, D.~R. Arrojo, and A.~J. Davison, ``{RLBench: The Robot
  Learning Benchmark \& Learning Environment},'' \emph{{IEEE} Robot. Autom.
  Lett.}, vol.~5, no.~2, pp. 3019--3026, 2020.

\bibitem{liu2023libero}
B.~Liu \emph{et~al.}, ``{LIBERO: Benchmarking Knowledge Transfer for Lifelong
  Robot Learning},'' in \emph{Adv. Neural Inf. Process. Syst. (NeurIPS)}, 2023.

\bibitem{mandlekar2021robomimic}
A.~Mandlekar \emph{et~al.}, ``{What Matters in Learning from Offline Human
  Demonstrations for Robot Manipulation},'' in \emph{Proc. Conf. Robot Learn.
  (CoRL)}, ser. PMLR, vol. 164, 2021, pp. 1678--1690.

\bibitem{nasiriany2024robocasa}
S.~Nasiriany \emph{et~al.}, ``{RoboCasa: Large-Scale Simulation of Household
  Tasks for Generalist Robots},'' in \emph{Proc. Robot. Sci. Syst. (RSS)},
  2024.

\bibitem{li2023behavior1k}
C.~Li \emph{et~al.}, ``{BEHAVIOR-1K: A Benchmark for Embodied AI with 1,000
  Everyday Activities and Realistic Simulation},'' in \emph{Proc. Conf. Robot
  Learn. (CoRL)}, ser. PMLR, vol. 205, 2022, pp. 80--93.

\bibitem{tao2024maniskill3}
S.~Tao \emph{et~al.}, ``{Demonstrating GPU Parallelized Robot Simulation and
  Rendering for Generalizable Embodied AI with ManiSkill3},'' in \emph{Proc.
  Robot. Sci. Syst. (RSS)}, 2025.

\bibitem{mandlekar2023mimicgen}
A.~Mandlekar \emph{et~al.}, ``{MimicGen: A Data Generation System for Scalable
  Robot Learning using Human Demonstrations},'' in \emph{Proc. Conf. Robot
  Learn. (CoRL)}, ser. PMLR, vol. 229, 2023, pp. 1820--1864.

\bibitem{jiang2025dexmimicgen}
Z.~Jiang \emph{et~al.}, ``{DexMimicGen: Automated Data Generation for Bimanual
  Dexterous Manipulation via Imitation Learning},'' in \emph{Proc. IEEE Int.
  Conf. Robot. Autom. (ICRA)}, 2025, pp. 16\,923--16\,930.

\bibitem{cheng2024exbody}
X.~Cheng, Y.~Ji, J.~Chen, R.~Yang, G.~Yang, and X.~Wang, ``{Expressive
  Whole-Body Control for Humanoid Robots},'' in \emph{Proc. Robot. Sci. Syst.
  (RSS)}, 2024.

\bibitem{ben2025homie}
Q.~Ben, F.~Jia, J.~Zeng, J.~Dong, D.~Lin, and J.~Pang, ``{HOMIE: Humanoid
  Loco-Manipulation with Isomorphic Exoskeleton Cockpit},'' in \emph{Proc.
  Robot. Sci. Syst. (RSS)}, 2025.

\bibitem{li2025amo}
J.~Li, X.~Cheng, T.~Huang, S.~Yang, R.-Z. Qiu, and X.~Wang, ``{AMO: Adaptive
  Motion Optimization for Hyper-Dexterous Humanoid Whole-Body Control},'' in
  \emph{Proc. Robot. Sci. Syst. (RSS)}, 2025.

\bibitem{ze2025twist}
Y.~Ze \emph{et~al.}, ``{TWIST: Teleoperated Whole-Body Imitation System},'' in
  \emph{Proc. Conf. Robot Learn. (CoRL)}, ser. PMLR, vol. 305, 2025, pp.
  2143--2154.

\bibitem{fu2024humanplus}
Z.~Fu, Q.~Zhao, Q.~Wu, G.~Wetzstein, and C.~Finn, ``{HumanPlus: Humanoid
  Shadowing and Imitation from Humans},'' in \emph{Proc. Conf. Robot Learn.
  (CoRL)}, ser. PMLR, vol. 270, 2024, pp. 2828--2844.

\bibitem{he2024omnih2o}
T.~He \emph{et~al.}, ``{OmniH2O: Universal and Dexterous Human-to-Humanoid
  Whole-Body Teleoperation and Learning},'' in \emph{Proc. Conf. Robot Learn.
  (CoRL)}, ser. PMLR, vol. 270, 2024, pp. 1516--1540.

\bibitem{cheng2024television}
X.~Cheng, J.~Li, S.~Yang, G.~Yang, and X.~Wang, ``{Open-TeleVision:
  Teleoperation with Immersive Active Visual Feedback},'' in \emph{Proc. Conf.
  Robot Learn. (CoRL)}, ser. PMLR, vol. 270, 2024, pp. 2729--2749.

\bibitem{sferrazza2024humanoidbench}
C.~Sferrazza, D.-M. Huang, X.~Lin, Y.~Lee, and P.~Abbeel, ``{HumanoidBench:
  Simulated Humanoid Benchmark for Whole-Body Locomotion and Manipulation},''
  in \emph{Proc. Robot. Sci. Syst. (RSS)}, 2024.

\bibitem{yang2025egovla}
R.~Yang \emph{et~al.}, ``{EgoVLA: Learning Vision-Language-Action Models from
  Egocentric Human Videos},'' \emph{arXiv preprint arXiv:2507.12440}, 2025.

\bibitem{fetchman2026}
O.~Rayyan \emph{et~al.}, ``{FetchMan: Learning Visual Humanoid
  Loco-Manipulation Policies from Simulated Experiences},'' \emph{arXiv
  preprint arXiv:2608.17027}, 2026.

\bibitem{humanoideveryday2025}
Z.~Zhao \emph{et~al.}, ``{Humanoid Everyday: A Comprehensive Robotic Dataset
  for Open-World Humanoid Manipulation},'' \emph{arXiv preprint
  arXiv:2510.08807}, 2025.

\bibitem{kim2024openvla}
M.~J. Kim \emph{et~al.}, ``{OpenVLA: An Open-Source Vision-Language-Action
  Model},'' in \emph{Proc. Conf. Robot Learn. (CoRL)}, ser. PMLR, vol. 270,
  2024, pp. 2679--2713.

\bibitem{pi05}
{Physical Intelligence} \emph{et~al.}, ``{$\pi_{0.5}$: a Vision-Language-Action
  Model with Open-World Generalization},'' \emph{arXiv preprint
  arXiv:2504.16054}, 2025.

\bibitem{nvidia2025groot}
{NVIDIA}, ``{GR00T N1: An Open Foundation Model for Generalist Humanoid
  Robots},'' \emph{arXiv preprint arXiv:2503.14734}, 2025.

\bibitem{zhao2023act}
T.~Z. Zhao, V.~Kumar, S.~Levine, and C.~Finn, ``{Learning Fine-Grained Bimanual
  Manipulation with Low-Cost Hardware},'' in \emph{Proc. Robot. Sci. Syst.
  (RSS)}, 2023.

\bibitem{chi2023diffusion}
C.~Chi \emph{et~al.}, ``{Diffusion Policy: Visuomotor Policy Learning via
  Action Diffusion},'' in \emph{Proc. Robot. Sci. Syst. (RSS)}, 2023.

\bibitem{seo2025cqnas}
Y.~Seo and P.~Abbeel, ``{Coarse-to-fine Q-Network with Action Sequence for
  Data-Efficient Reinforcement Learning},'' in \emph{Adv. Neural Inf. Process.
  Syst. (NeurIPS)}, 2025.

\bibitem{seo2024cqn}
Y.~Seo, J.~Uru{\c c}, and S.~James, ``{Continuous Control with Coarse-to-fine
  Reinforcement Learning},'' in \emph{Proc. Conf. Robot Learn. (CoRL)}, ser.
  PMLR, vol. 270, 2024, pp. 2866--2894.

\bibitem{yarats2022drqv2}
D.~Yarats, R.~Fergus, A.~Lazaric, and L.~Pinto, ``{Mastering Visual Continuous
  Control: Improved Data-Augmented Reinforcement Learning},'' in \emph{Proc.
  Int. Conf. Learn. Represent. (ICLR)}, 2022.

\bibitem{deas2025}
C.~Kim, H.~Lee, Y.~Seo, K.~Lee, and Y.~Zhu, ``{DEAS: DEtached value learning
  with Action Sequence for Scalable Offline RL},'' in \emph{Proc. Int. Conf.
  Learn. Represent. (ICLR)}, 2026.

\bibitem{todorov2012mujoco}
E.~Todorov, T.~Erez, and Y.~Tassa, ``{MuJoCo: A physics engine for model-based
  control},'' in \emph{Proc. IEEE/RSJ Int. Conf. Intell. Robots Syst. (IROS)},
  2012, pp. 5026--5033.

\bibitem{Zakka_Mink_Python_inverse_2026}
\BIBentryALTinterwordspacing
K.~Zakka, ``{Mink: Python inverse kinematics based on MuJoCo},'' Jun. 2026,
  version 1.2.0. [Online]. Available: \url{https://github.com/kevinzakka/mink}
\BIBentrySTDinterwordspacing

\bibitem{yi2025viser}
B.~Yi \emph{et~al.}, ``{Viser: Imperative, Web-based 3D Visualization in
  Python},'' \emph{arXiv preprint arXiv:2507.22885}, 2025.

\bibitem{henderson2018deeprl}
P.~Henderson, R.~Islam, P.~Bachman, J.~Pineau, D.~Precup, and D.~Meger, ``{Deep
  Reinforcement Learning That Matters},'' in \emph{Proc. AAAI Conf. Artif.
  Intell.}, vol.~32, no.~1, 2018.

\bibitem{agarwal2021rliable}
R.~Agarwal, M.~Schwarzer, P.~S. Castro, A.~Courville, and M.~G. Bellemare,
  ``{Deep Reinforcement Learning at the Edge of the Statistical Precipice},''
  in \emph{Adv. Neural Inf. Process. Syst. (NeurIPS)}, 2021.

\bibitem{colas2018seeds}
C.~Colas, O.~Sigaud, and P.-Y. Oudeyer, ``{How Many Random Seeds? Statistical
  Power Analysis in Deep Reinforcement Learning Experiments},'' \emph{arXiv
  preprint arXiv:1806.08295}, 2018.

\bibitem{snyder2025imitation}
D.~Snyder \emph{et~al.}, ``{Is Your Imitation Learning Policy Better than Mine?
  Policy Comparison with Near-Optimal Stopping},'' in \emph{Proc. Robot. Sci.
  Syst. (RSS)}, 2025.

\bibitem{ye2026dreamzero}
S.~Ye \emph{et~al.}, ``{World Action Models are Zero-shot Policies},''
  \emph{arXiv preprint arXiv:2602.15922}, 2026.

\end{thebibliography}

\clearpage
\onecolumn
\raggedbottom
\makeatletter
\setlength{\@fptop}{0pt}
\setlength{\@fpbot}{0pt plus 1fil}
\setlength{\@dblfptop}{0pt}
\setlength{\@dblfpbot}{0pt plus 1fil}
\setlength{\@dblfpsep}{14pt}
\makeatother
\renewcommand{\topfraction}{0.9}
\renewcommand{\bottomfraction}{0.8}
\renewcommand{\textfraction}{0.07}
\renewcommand{\floatpagefraction}{0.85}
\setcounter{topnumber}{4}
\setcounter{bottomnumber}{2}
\setcounter{totalnumber}{5}

\iflatexml\else\useRomanappendicesfalse\fi
\appendices

\renewcommand{\thetable}{A\arabic{table}}
\setcounter{table}{0}
\renewcommand{\thefigure}{A\arabic{figure}}
\setcounter{figure}{0}
\makeatletter
\renewcommand{\theHtable}{appendix.table.\arabic{table}}
\renewcommand{\theHfigure}{appendix.figure.\arabic{figure}}
\makeatother

\section{Tasks and Success Predicates}
\label{sec:app-tasks}
\label{app:tasks}

\begin{figure*}[!htbp]
\centering
\includegraphics[width=\textwidth]{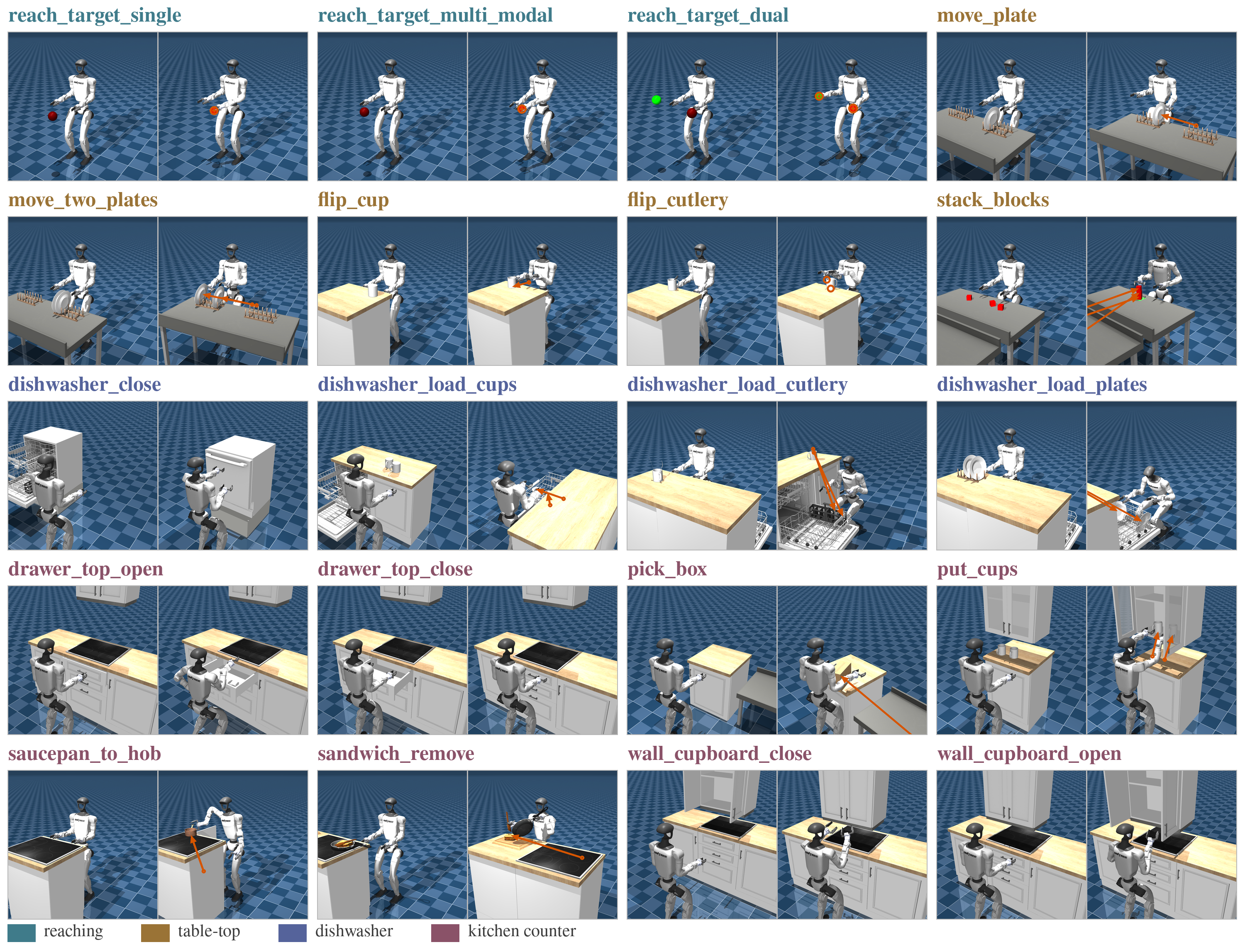}
\caption{\textbf{The \ntasks{} tasks of \benchname{}.} MuJoCo renders of one demonstration per task: the state after reset (left) and the final frame (right). The coloured rule above each pair marks the scene family, as in Fig.~\ref{fig:tasks}. The final frame marks the success criterion: the 5\,cm tolerance sphere for reaching, the object's recorded displacement for placement tasks and the measured joint extent against its threshold for articulated fixtures. Table~\ref{tab:tasks} gives the full predicates.}
\label{fig:task-catalog}
\end{figure*}

Fig.~\ref{fig:task-catalog} shows the \ntasks{} tasks in the four scene families of Section~\ref{sec:benchmark}, and Table~\ref{tab:tasks} gives each task's action dimension, episode budget and success predicate. Every predicate must hold for 1\,s (50 consecutive control steps). In the plate, cup, cutlery and block tasks an object touching the floor ends the episode as a failure; a robot fall is recorded but does not end the episode. The main comparison (Table~\ref{tab:main}) uses nine of them; Table~\ref{tab:coverage} reports ACT and CQN-AS on the other eleven.

\begin{table*}[!htbp]
\centering\small
\caption{\textbf{Episode budgets and success predicates of the \ntasks{} tasks.} Horizon is the episode budget in seconds (control steps at 50\,Hz). Every predicate must hold for 1\,s. Normalised joint travel is 0 at one end of a joint's range and 1 at the other. Tasks marked $^\dagger$ are outside the nine-task main comparison.}
\label{tab:tasks}
{\setlength{\tabcolsep}{4.5pt}\begin{tabular}{@{}lccp{9.6cm}@{}}
\toprule
\textbf{Task} & \textbf{Act.\ dim} & \textbf{Horizon} & \textbf{Success predicate (held for 1\,s)} \\
\midrule
\multicolumn{4}{@{}l}{\textit{Reaching (3 tasks)}} \\
\texttt{reach\_target\_single}$^\dagger$ & 20 & 18\,s (900) & Left end-effector (pinch-centre) site within 5\,cm of the target. \\
\texttt{reach\_target\_multi\_modal} & 20 & 14\,s (700) & Either end-effector site within 5\,cm of the target. \\
\texttt{reach\_target\_dual} & 20 & 14\,s (700) & Each end-effector site within 5\,cm of its own target, simultaneously. \\
\midrule
\multicolumn{4}{@{}l}{\textit{Tabletop (5 tasks)}} \\
\texttt{move\_plate} & 20 & 34\,s (1700) & Plate within 5\,cm of a target-rack slot, normal within $20^\circ$ of the slot axis, touching the rack and not the table, released. \\
\texttt{move\_two\_plates} & 21 & 46\,s (2300) & Both plates satisfy the single-plate predicate. \\
\texttt{flip\_cup}$^\dagger$ & 21 & 37\,s (1850) & Mug upright within $5^\circ$, on the counter, released. \\
\texttt{flip\_cutlery}$^\dagger$ & 21 & 39\,s (1950) & Spoon within $50^\circ$ of pointing down, touching the mug; spoon and mug released. \\
\texttt{stack\_blocks}$^\dagger$ & 21 & 175\,s (8750) & Bottom block on the pad, contact chain, each block $\ge 3$\,cm above and within 5\,cm planar offset of the one below, all released. \\
\midrule
\multicolumn{4}{@{}l}{\textit{Dishwasher (4 tasks)}} \\
\texttt{dishwasher\_close} & 21 & 93\,s (4650) & Door and both trays within 0.05 of closed (normalised joint travel). \\
\texttt{dishwasher\_load\_cups} & 21 & 40\,s (2000) & Both mugs touching the upper tray, released. \\
\texttt{dishwasher\_load\_cutlery}$^\dagger$ & 21 & 53\,s (2650) & Knife and fork in the cutlery basket, released. \\
\texttt{dishwasher\_load\_plates}$^\dagger$ & 21 & 63\,s (3150) & Both plates touching the lower tray, within $20^\circ$ of the slot axis, released. \\
\midrule
\multicolumn{4}{@{}l}{\textit{Kitchen counter (8 tasks)}} \\
\texttt{drawer\_top\_close} & 20 & 17\,s (850) & Top drawer within 0.1 of closed (normalised joint travel). \\
\texttt{drawer\_top\_open} & 20 & 27\,s (1350) & Top drawer within 0.1 of fully open (normalised joint travel). \\
\texttt{wall\_cupboard\_close}$^\dagger$ & 21 & 34\,s (1700) & Both doors within 0.1 of closed (normalised joint travel). \\
\texttt{wall\_cupboard\_open}$^\dagger$ & 21 & 29\,s (1450) & Both doors within 0.1 of fully open (normalised joint travel). \\
\texttt{pick\_box} & 21 & 88\,s (4400) & 3\,kg box from the 0.50\,m side table onto the 0.71\,m counter: in contact, bottom face within 3\,cm of the top, centre over the counter, released. \\
\texttt{put\_cups}$^\dagger$ & 21 & 60\,s (3000) & Both cups touching the bottom shelf of the wall cupboard, released. \\
\texttt{saucepan\_to\_hob}$^\dagger$ & 21 & 93\,s (4650) & Saucepan touching the hob, released. \\
\texttt{sandwich\_remove}$^\dagger$ & 21 & 63\,s (3150) & Sandwich touching the board with either face up within $10^\circ$; grasping the sandwich directly fails. \\
\bottomrule
\end{tabular}
}
\end{table*}

\textbf{Reset distributions and scene layout.}
Reset distributions are task-specific. Seventeen tasks randomise the scene per seed: object positions by 1--5\,cm and yaw by $\pm15^\circ$ to $\pm180^\circ$ depending on the task, reaching targets within a box around a nominal point, and, in the two drawer tasks, the robot pose ($x \in [0.12, 0.14]$\,m, $y \in [-0.05, 0.05]$\,m, yaw $\pm5^\circ$) and the initial drawer opening. Dishwasher closing and the two wall-cupboard tasks reset to a fixed state.

Furniture heights follow each scene. In box pickup the side table and the counter are 0.50\,m and 0.71\,m high; the default kitchen counter is about 0.85\,m; the two stacking tables are 0.70\,m high with centres 0.70\,m apart. The dishwasher-closing scene raises the dishwasher on a 0.25\,m plinth. The G1 is 1.32\,m tall.

\section{Robot, Controller and Policy Interfaces}
\label{sec:app-interfaces}
\label{app:interfaces}

\subsection{Embodiment Specifications and Whole-Body Execution}
\benchname{} executes all human demonstrations, learned policies and agent-written programs on a simulated Unitree G1 humanoid with 29 actuated degrees of freedom (12 in the legs, 3 at the waist and 14 in the two 7-DoF arms) and two \mbox{Dex1-1} parallel grippers. Frozen GR00T-WBC policies~\cite{nvidiawbc} receive the body commands at 50\,Hz and output position targets for the 12 leg joints and the 3 waist joints. MuJoCo position actuators with the controller's PD gains track these targets at the 1\,ms physics step, 20 substeps per control step.

\textbf{Passive pelvis tilt.}
The simulated base is a chain of planar slide and yaw joints, as in BiGym, extended with two passive, unactuated hinges for pelvis roll and pitch; every recorded demonstration uses them. With roll and pitch locked, a lateral command of $v_y=0.25$\,m/s produced $+0.164$\,m/s of lateral and $+0.059$\,m/s of unintended forward velocity. With the passive hinges, the response is $+0.1725$\,m/s lateral and $+0.014$\,m/s forward, against $+0.180$\,m/s lateral for the upstream free-floating reference.

\subsection{Action Representation and Controller Contract}
Table~\ref{tab:spaces} gives the 20- and 21-dimensional action layouts. The first four dimensions are body commands: forward velocity, lateral velocity, base height and yaw rate. The policy range $[-1,1]$ maps to the command ranges recorded in the demonstration metadata: $v_x \in [-0.35, 0.35]$\,m/s, $v_y \in [-0.25, 0.25]$\,m/s, $z_{\mathrm{base}} \in [0.40, 1.00]$\,m and $\omega_z \in [-0.50, 0.50]$\,rad/s. Fourteen tasks add a 21st dimension, an absolute torso-pitch reference $\theta_{\mathrm{pitch}} \in [-0.20, 0.80]$\,rad; the three reaching tasks, both drawer tasks and plate transport use 20 dimensions because their demonstrations were collected before this channel was added (Table~\ref{tab:tasks}).

The remaining channels are absolute joint targets for the two 7-DoF arms, normalised over each joint's range, and one closing command per \mbox{Dex1-1} gripper (0 fully open, 1 fully closed). All baselines except \pizero{} predict these absolute targets. \pizero{} is trained on relative arm actions, expressed against the arm state at the start of each predicted chunk, with base and gripper channels kept absolute (Appendix~\ref{app:vla}).

\subsection{Proprioceptive State and Multi-Camera Observations}
\begin{figure*}[!htbp]
\centering
\includegraphics[width=\textwidth]{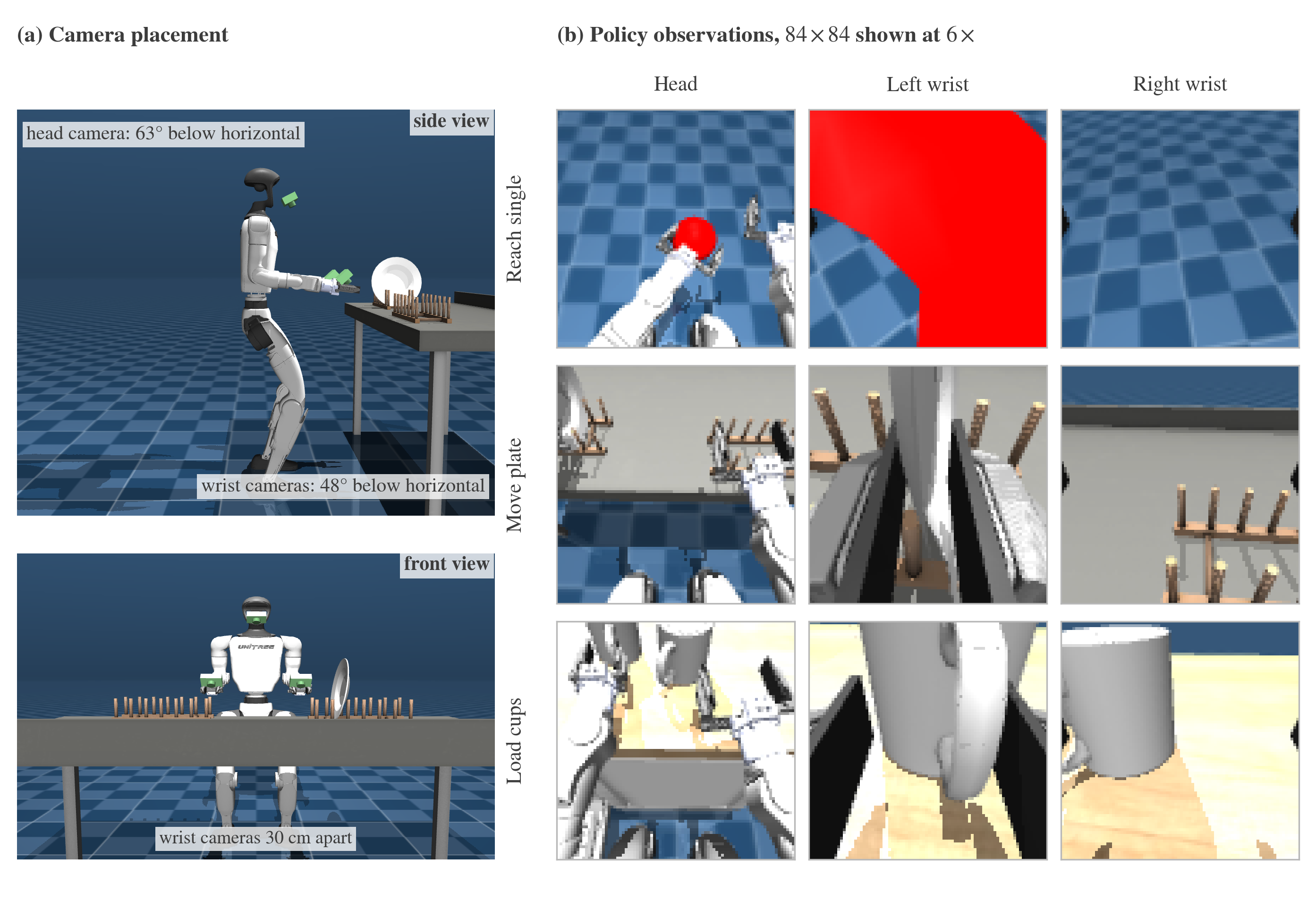}
\caption{\textbf{Onboard cameras.} (a) MuJoCo renders of the G1 at the start of a plate-transport demonstration; the green markers are the three cameras. The head camera sits on \texttt{torso\_link} and each wrist camera on its \texttt{wrist\_yaw\_link}, 4.9\,cm from the pinch centre between the finger pads. All three have a $60^\circ$ field of view, horizontally and vertically, because the image is square. (b) Stored $84\times84$ observations from one demonstration of each of three tasks, taken when a finger pad first touches an object (for reaching, at success), upscaled $6\times$ by nearest neighbour without other processing.}
\label{fig:onboard-cameras}
\end{figure*}

The non-VLA policies receive three synchronised $84\times84$ RGB views (Table~\ref{tab:spaces}, Fig.~\ref{fig:onboard-cameras}): a head camera on \texttt{torso\_link} and one camera on each \texttt{wrist\_yaw\_link}, all with $\mathrm{fovy}=60^\circ$. At the plate-transport reference pose the head camera is pitched $63^\circ$ below horizontal and the wrist cameras $48^\circ$, 30\,cm apart. Four consecutive frames are stacked per camera. Proprioception is 50-dimensional, or 56-dimensional on torso-pitch tasks: positions and velocities of the arm, finger and pelvis ($x$, $y$, $z$, yaw) joints, plus waist yaw, roll and pitch on 56-D tasks, followed by two gripper-closure scalars and the pelvis pose. It is stored raw and standardised with demonstration statistics when loaded.

\textbf{Reset determinism.}
The GR00T-WBC adapter keeps internal state across control steps, so a seeded reset alone does not reproduce a trajectory. With evaluation-time controller restoration disabled, a seeded reset of an environment that has already run a 60-step episode differs from the same reset of a fresh environment by $9.5\times10^{-2}$ in configuration (maximum absolute \texttt{qpos} difference), growing to $2.1\times10^{-1}$ within 50 steps. \benchname{} therefore restores, at every evaluation reset, the controller state captured at the first reset. Fig.~\ref{fig:replay-scatter} shows the effect on open-loop replay of one plate-transport demonstration: 20 replays from the recorded engage snapshot (simulator and controller state) are identical at every step, whereas after a seed-only reset the final plate positions differ by up to 0.73\,mm. All 40 replays succeed.

\begin{table*}[!htbp]
\centering\small
\caption{\textbf{Action and observation layout of the G1.} The standard configuration has 20 action and 50 proprioceptive dimensions; torso-pitch tasks ($^*$) have 21 and 56. Command ranges are those recorded in the demonstration metadata.}
\label{tab:spaces}
{\setlength{\tabcolsep}{4pt}\resizebox{\textwidth}{!}{%
\begin{tabular}{@{}llp{5.6cm}ll@{}}
\toprule
\textbf{Channel} & \textbf{Slice} & \textbf{Content} & \textbf{Physical range} & \textbf{Policy range} \\
\midrule
\multicolumn{5}{@{}l}{\textit{Action (20-D; 21-D with torso pitch$^*$), 50\,Hz}} \\
Forward velocity $v_x$ & $[0]$ & Body command to GR00T-WBC & $[-0.35, 0.35]$\,m/s & $[-1, 1]$ \\
Lateral velocity $v_y$ & $[1]$ & Body command to GR00T-WBC & $[-0.25, 0.25]$\,m/s & $[-1, 1]$ \\
Base height $z_{\mathrm{base}}$ & $[2]$ & Body command to GR00T-WBC & $[0.40, 1.00]$\,m & $[-1, 1]$ \\
Yaw rate $\omega_z$ & $[3]$ & Body command to GR00T-WBC & $[-0.50, 0.50]$\,rad/s & $[-1, 1]$ \\
Torso pitch $\theta_{\mathrm{pitch}}$$^*$ & $[4]^*$ & Waist pitch reference & $[-0.20, 0.80]$\,rad & $[-1, 1]$ \\
Left arm & $[4{:}11]$ / $[5{:}12]^*$ & Absolute joint targets: shoulder P/R/Y, elbow, wrist R/P/Y & Joint range & $[-1, 1]$ \\
Right arm & $[11{:}18]$ / $[12{:}19]^*$ & Absolute joint targets, same order & Joint range & $[-1, 1]$ \\
Grippers & $[18{:}20]$ / $[19{:}21]^*$ & Left, right \mbox{Dex1-1} command & 0 open, 1 closed & $[-1, 1]$ \\
\midrule
\multicolumn{5}{@{}l}{\textit{Proprioception (50-D; 56-D$^*$), stored raw, standardised with demonstration statistics at load time}} \\
Joint positions $q$ & $[0{:}22]$ / $[0{:}25]^*$ & (Waist Y/R/P$^*$,) left arm 7, left fingers 2, right arm 7, right fingers 2, pelvis $x, y, z$, yaw & rad, m & Standardised \\
Joint velocities $\dot q$ & $[22{:}44]$ / $[25{:}50]^*$ & Same joints as $q$ & rad/s, m/s & Standardised \\
Gripper state & $[44{:}46]$ / $[50{:}52]^*$ & Left, right closure & 0 open, 1 closed & Standardised \\
Pelvis pose & $[46{:}50]$ / $[52{:}56]^*$ & Pelvis $x, y, z$, yaw (simulator state) & m, rad & Standardised \\
\midrule
\multicolumn{5}{@{}l}{\textit{Images: three onboard RGB cameras, $84\times84$, four stacked frames}} \\
Head & View 1 & On \texttt{torso\_link}, $\mathrm{fovy}=60^\circ$ & uint8 & Method-specific \\
Left wrist & View 2 & On left \texttt{wrist\_yaw\_link}, $\mathrm{fovy}=60^\circ$ & uint8 & Method-specific \\
Right wrist & View 3 & On right \texttt{wrist\_yaw\_link}, $\mathrm{fovy}=60^\circ$ & uint8 & Method-specific \\
\bottomrule
\end{tabular}%
}
}
\end{table*}

\subsection{Simulation Throughput and Training Compute}
\label{sec:app-throughput}
We timed a single environment built by the benchmark protocol constructor in one process, with 1\,ms physics and 20 substeps per 50\,Hz control step, on an otherwise idle NVIDIA RTX 5090 of a workstation with an AMD Ryzen Threadripper PRO 7975WX (32 cores), using MuJoCo 3.8.1 with EGL rendering. The benchmark process was pinned to one CPU core while other training jobs ran on the remaining cores (load average 16 on 64 threads). Each configuration was stepped for 100 warm-up steps and then three repeats of 1,000 steps, either holding a fixed action or replaying demonstration actions; resets were timed separately. With GR00T-WBC in the loop and rendering disabled, the environment runs at 360--550 control steps/s, 7--11 times real time; with the benchmark observation of three $84\times84$ cameras it runs at 210--278 steps/s, 4--6 times real time (Table~\ref{tab:throughput}). Rendering the same cameras at $224\times224$ costs at most a further 8\%. A reset, including the 200-step controller settle, takes 0.36--0.47\,s.

Training to 101k updates or environment steps took, per run, a median of 3.3\,h for CQN-AS, 5.9\,h for ACT (12.4\,h when two runs shared a GPU), 7.9\,h for DP, 5.3\,h for DEAS and 1.7\,h for DrQ-v2+, each on one RTX 5090 of the same machine. These times exclude the independent checkpoint evaluation and depend on how the machine was shared.

\begin{table}[!htbp]
\centering\footnotesize
\caption{\textbf{Single-environment throughput in control steps per second} (mean $\pm$ std over three repeats of 1,000 steps; hold action / demonstration replay) and reset time (mean $\pm$ std, 30 resets). Measured on an idle RTX 5090, with the process pinned to one core of a Threadripper PRO 7975WX.}
\label{tab:throughput}
{\setlength{\tabcolsep}{2.5pt}%
\begin{tabular}{@{}lcccc@{}}
\toprule
 & \multicolumn{3}{c}{Control steps/s} & Reset \\
\cmidrule(lr){2-4}
Task & No render & $3\times84^2$ & $3\times224^2$ & (s) \\
\midrule
Move plate & 399 / 360 & 223 / 210 & 211 / 196 & 0.47 \\
Reach dual & 528 / 550 & 278 / 268 & 256 / 270 & 0.36 \\
Dishwasher close & 457 / 405 & 232 / 210 & 225 / 204 & 0.43 \\
\bottomrule
\end{tabular}}
\end{table}

\FloatBarrier
\section{Human Demonstrations and Dataset Format}
\label{sec:app-data}
\label{app:data}

\subsection{VR Teleoperation}
Operators wore a Meta Quest 3 headset connected through WiVRn. The headset shows a stereo MuJoCo render placed at the robot's head camera, with the head shell hidden; this view is separate from the $84\times84$ policy camera. Mink quadratic-programming inverse kinematics~\cite{Zakka_Mink_Python_inverse_2026} converts each tracked hand pose into seven arm-joint targets. The left thumbstick commands planar velocity; the right thumbstick commands yaw rate horizontally and base height vertically, the latter integrated over time. On torso-pitch tasks a left-stick click toggles a torso-pitch mode, and the triggers close the grippers. The environment steps at 50\,Hz independently of the headset frame rate. The same frozen GR00T-WBC executes the body commands during collection and evaluation, so the demonstrations contain the controller's tracking lag and gait-induced torso motion.

\subsection{Demonstration Motion Statistics}
\begin{figure*}[!htbp]
\centering
\includegraphics[width=\textwidth]{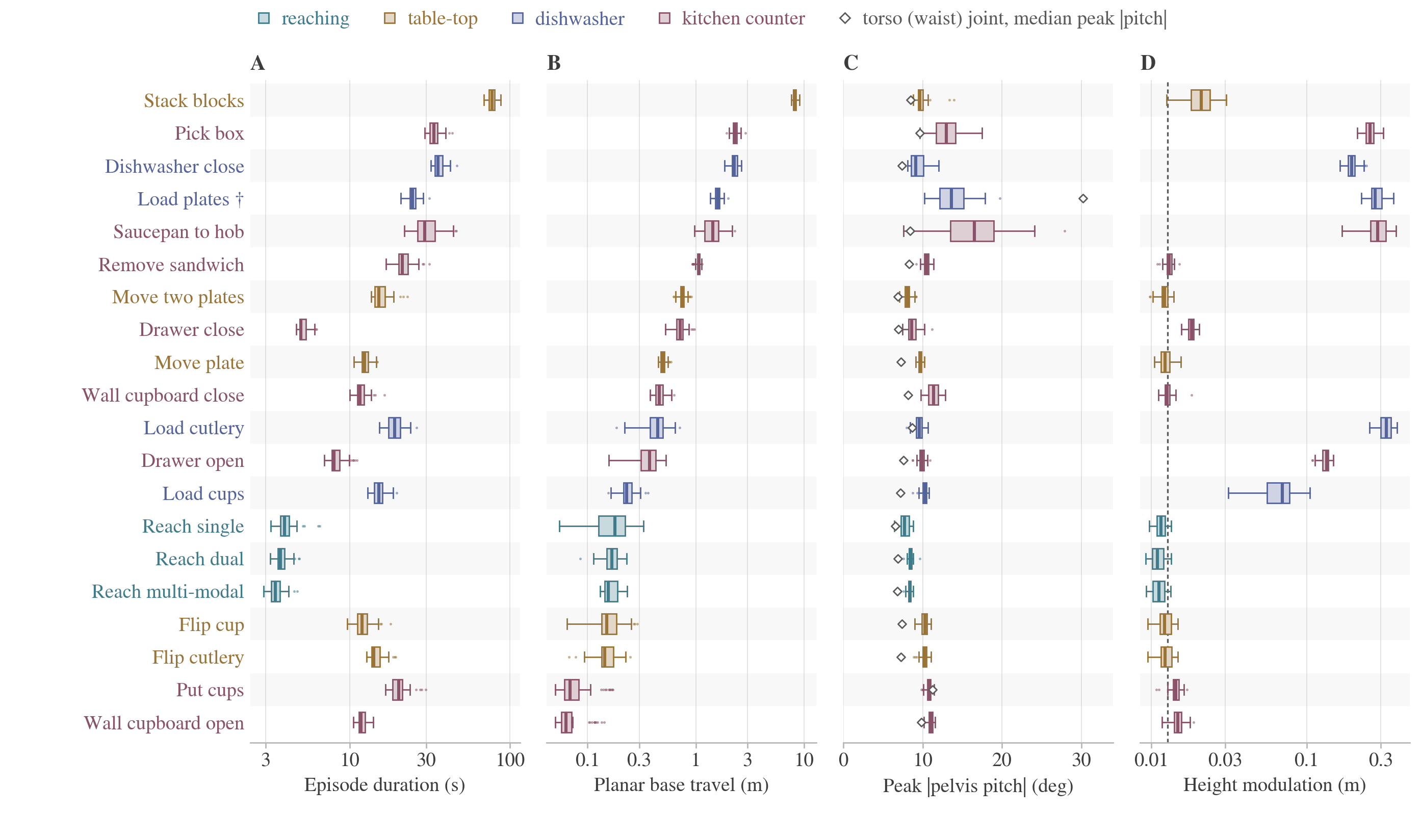}
\caption{\textbf{Kinematic distributions of the 1{,}200 human VR demonstrations}: \ntasks{} tasks $\times$ \ndemos{} successful episodes. \textbf{A}: episode duration (median 14.4\,s). \textbf{B}: planar base travel, from 0.06\,m near-stationary manipulation to 8.10\,m cross-workspace transport. \textbf{C}: peak absolute pelvis lean within an episode, not a net or mean lean. \textbf{D}: physical pelvis height modulation, which is the executed travel rather than the command; the dashed line marks the 1.3\,cm gait floor of the 11 tasks whose height command never moves. Boxes give the median and interquartile range, whiskers extend to $1.5\times$ IQR and dots are episodes beyond them. Tasks are ordered by median base travel, so one row reads across all four panels; A, B and D are log-scaled. $^\dagger$Plate loading is the only task whose commanded torso-pitch action ever leaves 0\,rad (58/60 episodes, peak 0.8\,rad), yet every episode of all \ntasks{} tasks leans $6.4$--$27.9^\circ$.}
\label{fig:demo-stats}
\end{figure*}

Fig.~\ref{fig:demo-stats} summarises the 1,200 demonstrations:
\begin{itemize}
  \item \textbf{Duration}: the median episode lasts 14.4\,s (range 2.9--87.4\,s); per-task medians range from 3.4\,s (\texttt{reach\_target\_multi\_modal}) to 77.1\,s (\texttt{stack\_blocks}). The episode budgets in Table~\ref{tab:tasks} are longer.
  \item \textbf{Base travel}: per-task median planar base travel ranges from 0.06\,m (\texttt{wall\_cupboard\_open}) to 8.10\,m (\texttt{stack\_blocks}), where every demonstration makes two outbound carries and one return.
  \item \textbf{Height and posture}: pelvis height spans 0.38--0.78\,m across episodes (commands 0.40--0.80\,m); 11 tasks never change the height command. Only \texttt{dishwasher\_load\_plates} commands torso pitch (58 of 60 episodes, up to 0.8\,rad), yet every episode of every task reaches a peak pelvis lean of $6.4^\circ$--$27.9^\circ$.
\end{itemize}

\subsection{Recording Format and Replay}
\begin{figure}[!htbp]
\centering
\iflatexml
\includegraphics[width=0.74\textwidth]{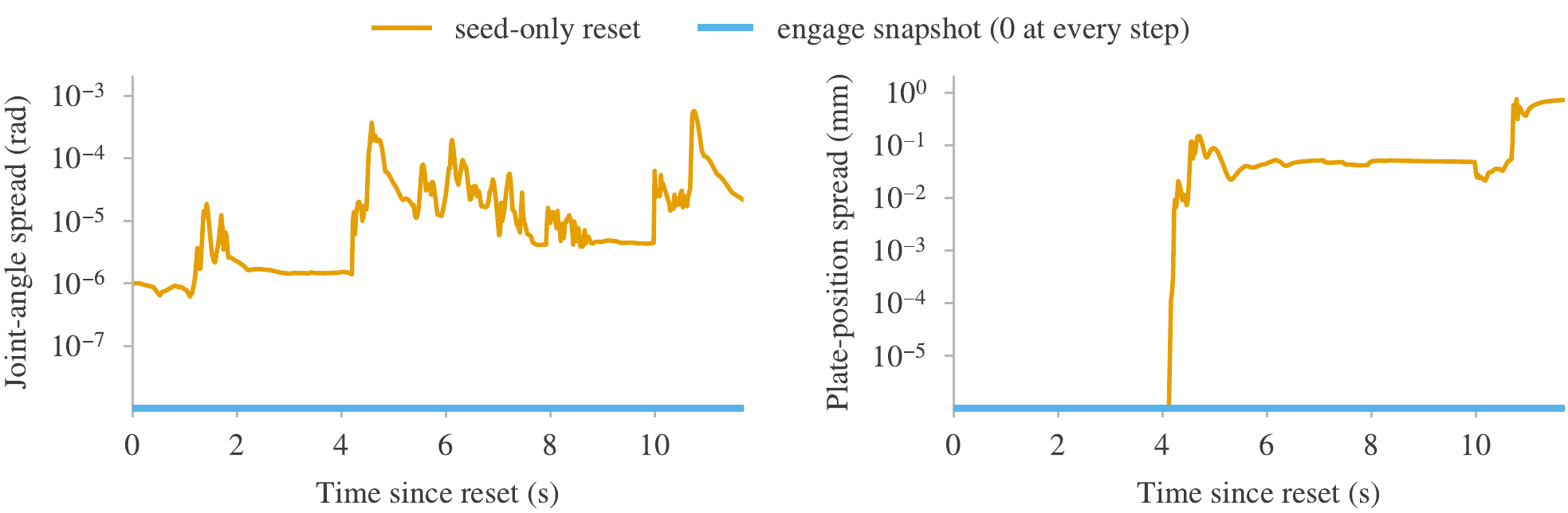}
\else
\includegraphics[width=0.74\textwidth]{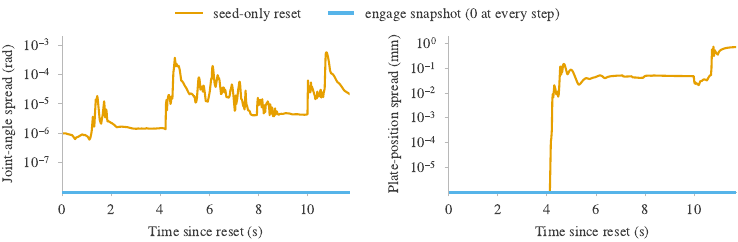}
\fi
\caption{\textbf{Replay determinism.} One \texttt{move\_plate} demonstration (seed 18) replayed open-loop 20 times from its recorded engage snapshot and 20 times after a seed-only reset. Each curve is the largest difference to the first replay of the same condition, over the 29 actuated joint angles (left) and the plate position (right). From the snapshot, the replays are identical at every step. After a seed-only reset, the joint angles start $10^{-6}$\,rad apart and the difference grows along the trajectory; the plate positions coincide until the gripper reaches the plate at about 4\,s and end at most 0.73\,mm apart. All 40 replays succeed.}
\label{fig:replay-scatter}
\end{figure}

The release contains \ndemos{} successful demonstrations for each of the \ntasks{} tasks, 1,200 episodes, each stored as a compressed NumPy \texttt{.npz} archive with JSON metadata. Per-step arrays include:
\begin{itemize}
  \item \texttt{rgb\_obs}: uint8 images of shape $(T, 3, 3, 84, 84)$, one per camera, in the order recorded in the metadata;
  \item \texttt{low\_dim\_obs}: raw 50- or 56-dimensional float32 proprioception;
  \item \texttt{action}: the normalised 20- or 21-dimensional policy action, with physical values in \texttt{raw\_outer\_action};
  \item \texttt{reward} and \texttt{discount};
  \item \texttt{full\_qpos} and \texttt{full\_qvel}: the complete simulator configuration and velocity;
  \item \texttt{lowerbody\_command}, \texttt{height\_command}, \texttt{torso\_target} and \texttt{leg\_joint\_targets}: controller inputs and outputs.
\end{itemize}
Each episode also stores its seed and an engage snapshot taken when the operator takes control: \texttt{init\_qpos}, \texttt{init\_qvel}, \texttt{init\_ctrl}, \texttt{init\_qacc\_warmstart} and the controller state (\texttt{lb\_state.*}). Replaying the recorded actions from this snapshot reproduces the recorded trajectory; Fig.~\ref{fig:replay-scatter} shows 20 identical replays of one demonstration. Collection requires a 3\,s success hold; the training view keeps the first 1\,s of it, and the hold length is recorded in the metadata.

\FloatBarrier
\section{Baseline Implementation and Training Recipes}
\label{sec:app-baselines}
\label{app:baselines}

Table~\ref{tab:hyperparams} lists the settings resolved from the run configurations of the main-table experiments; they were identical across the runs of each method.

\subsection{Imitation Learning: ACT and Diffusion Policy}
\textbf{Action Chunking with Transformers (ACT).} We use our JAX port of ACT~\cite{zhao2023act}. Each camera view is encoded by an ImageNet-pretrained ResNet-18 with frozen batch normalisation; image tokens and a linear proprioception token enter a transformer encoder--decoder (4 encoder layers, 1 decoder layer, 8 heads, width 512) with a conditional VAE (32-dimensional latent, $\beta_{\mathrm{KL}}=10$). ACT predicts $K=32$ actions, replans every step and fuses overlapping predictions by temporal ensembling with weights $\exp(-0.01\,i)$. Training uses AdamW with learning rate $10^{-4}$, $10^{-5}$ for the image encoder, weight decay $10^{-4}$ and batch 256 for 101k updates.

\textbf{Diffusion Policy (DP).} Each view is encoded by an ImageNet-pretrained ResNet-18 with frozen batch normalisation and spatial softmax. The image and proprioception features condition a 1D temporal U-Net (channels 256, 512, 1024) through FiLM. DP uses 100 DDPM steps in training and 10 DDIM steps at inference, predicts $K=16$ actions and executes 8 before replanning. Training uses AdamW with learning rate $10^{-4}$, $10^{-5}$ for the image encoder, weight decay $10^{-6}$, gradient clipping at 1.0, an exponential moving average of the weights (0.9999, used for evaluation) and batch 256 for 101k updates.

\subsection{Demo-Driven Reinforcement Learning: CQN-AS, DrQ-v2+ and DEAS}
\textbf{CQN-AS.} CQN-AS~\cite{seo2025cqnas} is a critic-only method over action sequences. Each camera has its own 4-layer CNN, and a GRU encodes the action sequence. Actions are discretised coarse to fine (3 levels of 5 bins per dimension), and the critic is a C51 distribution over 51 atoms on $[-2, 2]$. The policy predicts $K=32$ actions, replans every step and uses the same temporal ensembling as ACT. Each update samples 256 online and 256 demonstration transitions. The loss weights the distributional TD term by 0.1 and the demonstration terms, a first-order stochastic dominance term and a margin loss (margin 0.1), by 0.9. Training uses AdamW with learning rate $5\times10^{-5}$ and weight decay 0.1 over 101k environment steps, one update per step. Collection actions receive Gaussian noise of standard deviation 0.01 on all dimensions plus 0.03 on the first three body-command dimensions, and successful online episodes can enter the demonstration buffer.

\textbf{DrQ-v2+.} DrQ-v2+~\cite{yarats2022drqv2,seo2024cqn} is a single-step ($K=1$) actor--critic with a deterministic actor and twin distributional critics (101-bin categorical on $[-2, 2]$, hidden width 1024). Each camera has its own 4-layer CNN, trained through the critic loss. It uses the same 256+256 sampling and collection noise as CQN-AS, an MSE behaviour-cloning term with weight 1.0, and AdamW with learning rate $10^{-4}$ and weight decay 0.1.

\textbf{DEAS.} DEAS~\cite{deas2025} learns offline from the demonstrations. Twin critics use an HL-Gauss distribution (101 bins on $[0, 1]$); a distributional value function is fitted by expectile regression ($\tau=0.7$); a deterministic actor predicts $K=16$ actions, all executed, and is extracted by advantage-weighted regression ($\beta=1$, weights clipped at 100). The discount is 0.9 within a chunk and 0.99 for bootstrapping. Each camera has its own 4-layer CNN. Training uses Adam with learning rate $10^{-4}$, no weight decay and batch 256 for 101k updates.

\subsection{Vision-Language-Action (VLA) Fine-Tuning: \texorpdfstring{\pizero{}}{pi0.5}}
\label{sec:app-vla-recipe}
\label{app:vla}
We fine-tune the released \pizero{} base checkpoint independently on each task with full supervised fine-tuning (Table~\ref{tab:vla-hparams}). The model combines a PaliGemma vision-language backbone with a SigLIP-So400M image encoder and a flow-matching action expert. The task instruction (Table~\ref{tab:vla-instructions}) and the proprioceptive state, discretised into 256 bins, enter as text tokens. \pizero{} predicts a 50-step action chunk and executes the first 16 steps (0.32\,s at 50\,Hz) before replanning. Arm actions are relative to the arm state at the start of the chunk; base and gripper actions are absolute. Each task trains for 30k updates.

\subsection{Shared Training Augmentation}
All non-VLA baselines use the same training augmentation. Images are randomly shifted by up to 8 pixels, padding by edge replication, independently per sample and view. With probability 0.2 per sample, the entire non-base proprioceptive vector is set to zero across all stacked frames; after standardisation, zero is the demonstration mean. The pelvis entries are kept. Neither augmentation is applied at evaluation.

\begin{table*}[!htbp]
\centering\small
\caption{\textbf{Hyperparameters of the evaluated methods.} Non-VLA values are resolved from the run configurations of the main-table experiments. The online methods (CQN-AS, DrQ-v2+) sample 256 replay and 256 demonstration transitions per update and perform one update per environment step.}
\label{tab:hyperparams}
{\setlength{\tabcolsep}{3pt}{\scriptsize\renewcommand{\arraystretch}{1.0}%
\begin{tabular}{@{}>{\raggedright\arraybackslash}p{2.5cm}*{6}{>{\centering\arraybackslash}p{2.28cm}}@{}}
\toprule
\textbf{Hyperparameter} & \textbf{ACT} & \textbf{DP} & \textbf{CQN-AS} & \textbf{DrQ-v2+} & \textbf{DEAS} & \textbf{\pizero{}} \\
\midrule
Policy representation & Transformer CVAE & 1D temporal U-Net (FiLM) & C2F C51 critic, GRU over chunk & Actor + distributional twin critic & Distributional critics + AWR actor & PaliGemma VLM + flow-matching expert \\
Visual encoder & ResNet-18 (ImageNet, frozen BN) & ResNet-18 (ImageNet, frozen BN), spatial softmax & 4-layer CNN per camera & 4-layer CNN per camera & 4-layer CNN per camera & SigLIP-So400M \\
Proprioception input & Linear token & Projected, concatenated & Projected, concatenated & Projected, concatenated & Projected, concatenated & 256-bin discretised text tokens \\
Action chunk ($K$) & 32 & 16 & 32 & 1 & 16 & 50 \\
Executed per prediction ($H$) & 1 & 8 & 1 & 1 & 16 & 16 \\
Temporal ensemble & Yes ($m=0.01$) & No & Yes ($m=0.01$) & No & No & No \\
Action representation & Absolute & Absolute & Absolute, discretised (3 levels $\times$ 5 bins) & Absolute & Absolute & Arm delta; base, gripper absolute \\
Optimiser & AdamW & AdamW & AdamW & AdamW & Adam & AdamW \\
Learning rate (policy) & $10^{-4}$ & $10^{-4}$ & $5\times10^{-5}$ & $10^{-4}$ & $10^{-4}$ & $2.5\times10^{-5}$ \\
Learning rate (visual encoder) & $10^{-5}$ & $10^{-5}$ & $5\times10^{-5}$ (same) & $10^{-4}$ (same) & $10^{-4}$ (same) & $2.5\times10^{-5}$ (same) \\
Learning-rate schedule & Constant & Constant & Constant & Constant & Constant & 1k warmup, cosine to $2.5\times10^{-6}$ \\
Weight decay & $10^{-4}$ & $10^{-6}$ & 0.1 & 0.1 & 0 & $10^{-2}$ \\
EMA / target network & -- & EMA 0.9999, grad.\ clip 1.0 & Target $\tau=0.02$ & Hard copy every 100 updates & Target $\tau=0.005$ & -- \\
Batch size & 256 & 256 & 256 replay + 256 demo & 256 replay + 256 demo & 256 & 48 \\
Frame stack & 4 & 4 & 4 & 4 & 4 & 1 \\
Training augmentation & \multicolumn{5}{>{\centering\arraybackslash}p{12.2cm}}{Shift (pad 8, replicate); per-sample non-base proprioception mask, $p=0.2$} & None \\
Exploration noise & -- & -- & Gaussian 0.01 + base 0.03 & Gaussian 0.01 + base 0.03 & -- & -- \\
Loss & L1 + KL ($\beta=10$) & $\epsilon$-prediction MSE (100 DDPM train / 10 DDIM steps) & $0.1\,$C51 TD $+\,0.9\,$(FOSD + margin 0.1) & Distributional TD + MSE BC ($\lambda=1$) & HL-Gauss TD + expectile ($\tau=0.7$) + AWR ($\beta=1$) & Flow matching (MSE) \\
Training budget & 101k updates & 101k updates & 101k env.\ steps & 101k env.\ steps & 101k updates & 30k updates \\
\bottomrule
\end{tabular}%
}
}
\end{table*}

\begin{table*}[!htbp]
\centering\small
\caption{\textbf{\pizero{} fine-tuning settings.} One training run per task.}
\label{tab:vla-hparams}
{\setlength{\tabcolsep}{6pt}\begin{tabular}{@{}>{\raggedright\arraybackslash}p{0.22\textwidth}>{\raggedright\arraybackslash}p{0.70\textwidth}@{}}
\toprule
Setting & Value \\
\midrule
Base checkpoint & Released \pizero{} (\texttt{lerobot/pi05\_base}) \\
Adaptation & Full supervised fine-tuning (no LoRA) \\
Demonstrations & \ndemos{} successful VR episodes per task \\
Cameras & Head, left wrist, right wrist; $84\times84$ RGB (model $224\times224$) \\
Proprioception & 50-D (no torso pitch) or 56-D (pitch tasks) \\
Action & Relative arm delta; base and gripper absolute \\
Action dimension & 20 (no pitch) or 21 (with torso pitch) \\
Chunk / executed & 50 / 16 outer steps (0.32\,s at 50\,Hz) \\
Optimiser & AdamW, peak lr $2.5\times10^{-5}$, cosine to $2.5\times10^{-6}$, 1k warmup \\
Batch & 48 (global) \\
Precision & bfloat16 \\
Train seed & 1000 (one run per task) \\
Training horizon & 30k updates for every task \\
\bottomrule
\end{tabular}
}
\end{table*}

\begin{table*}[!htbp]
\centering\small
\caption{\textbf{Language instructions} used during \pizero{} fine-tuning and evaluation. $^\dagger$Not in the nine-task main comparison.}
\label{tab:vla-instructions}
{\setlength{\tabcolsep}{6pt}\begin{tabular}{@{}l>{\raggedright\arraybackslash}p{0.72\textwidth}@{}}
\toprule
Task & Instruction \\
\midrule
Pick box & Pick up the box from the side table and place it on the counter. \\
Reach multi-modal & Reach the target with either wrist. \\
Reach dual & Reach the two targets, one with each wrist. \\
Drawer close & Close the top drawer of the kitchen cabinet. \\
Drawer open & Open the top drawer of the kitchen cabinet. \\
Move plate & Move the plate between two draining racks. \\
Move two plates & Move two plates simultaneously from one draining rack to the other. \\
Dishwasher close & Push back all trays and close the door of the dishwasher. \\
Dishwasher load cups & Move the cups from the table into the dishwasher's upper tray. \\
Reach single$^\dagger$ & Reach the target with the left wrist. \\
\bottomrule
\end{tabular}
}
\end{table*}

\FloatBarrier
\section{Evaluation Protocol and Checkpoint Selection}
\label{sec:app-analysis}
\label{app:analysis}

Each evaluated checkpoint is rolled out for 100 episodes with seeds $\seedbase + i$, $i \in \{0, \dots, 99\}$, disjoint from the demonstration seeds (1--199). The score in Table~\ref{tab:main} is the \emph{last-five mean}: for each training run, the mean success over its last five saved checkpoints in the nominal 80k--100k window, 500 episodes per run. We report the mean and standard error across the three runs. \pizero{} has one training run per task; its score pools 50 episodes at each of its last five checkpoints (26k--30k) and has no across-run error bar.

\textbf{Peak checkpoint versus last-five mean.}
Table~\ref{tab:appendix-peak-final} compares the last-five mean with the mean over runs of each run's best checkpoint. With 100 episodes per checkpoint, the maximum over about 20 checkpoints is biased upwards, and choosing it consults the evaluation seeds. The gap $\Delta$ also contains genuine late-training decline, as on dishwasher closing, where DEAS runs peak at 97.0\% on average and end at 27.2\%. It is smallest for DP ($+5.5$ points on average) and largest for DEAS ($+13.6$). We report the last-five mean because it is fixed in advance.

\begin{table*}[!htbp]
\centering\small
\caption{\textbf{Peak checkpoint versus last-five mean on the nine main tasks.} Peak: mean over three runs of each run's best checkpoint (100 episodes per checkpoint). Last-5: the main-table score. $\Delta = \text{Peak} - \text{Last-5}$ in percentage points. DrQ-v2+ is omitted (99\% last-five mean on drawer closing, 0\% on the other eight tasks).}
\label{tab:appendix-peak-final}
{\setlength{\tabcolsep}{3.5pt}\resizebox{\textwidth}{!}{%
\begin{tabular}{@{}l rrr rrr rrr rrr@{}}
\toprule
 & \multicolumn{3}{c}{\textbf{ACT}} & \multicolumn{3}{c}{\textbf{DP}} & \multicolumn{3}{c}{\textbf{CQN-AS}} & \multicolumn{3}{c}{\textbf{DEAS}} \\
\cmidrule(lr){2-4} \cmidrule(lr){5-7} \cmidrule(lr){8-10} \cmidrule(lr){11-13}
\textbf{Task} & Peak & Last-5 & $\Delta$ & Peak & Last-5 & $\Delta$ & Peak & Last-5 & $\Delta$ & Peak & Last-5 & $\Delta$ \\
\midrule
Pick box & 71.7 & 61.3 & +10.4 & 18.3 & 9.9 & +8.4 & 55.0 & 47.6 & +7.4 & 26.7 & 13.7 & +13.0 \\
Reach multi-modal & 83.3 & 80.6 & +2.7 & 87.0 & 83.7 & +3.3 & 28.3 & 18.9 & +9.5 & 85.7 & 79.3 & +6.4 \\
Reach dual & 33.0 & 29.9 & +3.1 & 30.7 & 26.8 & +3.9 & 10.7 & 2.9 & +7.8 & 40.0 & 33.8 & +6.2 \\
Drawer close & 100.0 & 99.8 & +0.2 & 100.0 & 100.0 & +0.0 & 100.0 & 99.9 & +0.1 & 100.0 & 100.0 & +0.0 \\
Drawer open & 94.3 & 92.3 & +2.1 & 99.3 & 98.1 & +1.2 & 93.3 & 89.1 & +4.2 & 90.7 & 78.7 & +12.0 \\
Move plate & 70.0 & 60.1 & +9.9 & 67.3 & 61.7 & +5.7 & 53.3 & 48.2 & +5.1 & 34.7 & 26.8 & +7.9 \\
Move two plates & 46.0 & 38.7 & +7.3 & 23.3 & 18.5 & +4.9 & 30.7 & 21.1 & +9.5 & 4.0 & 2.1 & +1.9 \\
Dishwasher close & 100.0 & 93.3 & +6.7 & 60.3 & 50.9 & +9.4 & 100.0 & 78.1 & +21.9 & 97.0 & 27.2 & +69.8 \\
Dishwasher load cups & 68.0 & 55.6 & +12.4 & 63.7 & 51.3 & +12.4 & 59.3 & 46.9 & +12.5 & 11.0 & 5.9 & +5.1 \\
\midrule
\textbf{Mean} & \textbf{74.0} & \textbf{67.9} & \textbf{+6.1} & \textbf{61.1} & \textbf{55.7} & \textbf{+5.5} & \textbf{59.0} & \textbf{50.3} & \textbf{+8.7} & \textbf{54.4} & \textbf{40.8} & \textbf{+13.6} \\
\bottomrule
\end{tabular}%
}
}
\end{table*}

\FloatBarrier
\section{Contact Modelling and a DrQ-v2+ Failure Mode}
\label{sec:app-qualitative}
\label{app:qualitative}

\subsection{Gait-Induced Grip Slip}
Walking produces periodic foot-contact impacts; in one replayed plate-transport demonstration, foot-force peaks recur at 3.45\,Hz (median interval 0.29\,s). Early plate-transport runs showed grasped plates slowly rotating in the gripper. We traced this to torsional friction at the \mbox{Dex1-1} finger pads, which have sliding friction 1.5 and contact priority 1, so their parameters govern pad--plate contacts. In a reduced model with the gait disturbance amplified threefold, 6\,s of transport accumulated 0.745\,mm of slip with the original pad torsional friction of 0.01:
\begin{equation}
\mathbf{0.745\,\mathrm{mm}} \xrightarrow{\mu_{\mathrm{torsion}}=0.05} \mathbf{0.002\,\mathrm{mm}} \xrightarrow{\mathrm{solref/solimp}} \mathbf{0.000\,\mathrm{mm}}
\end{equation}
Raising pad torsional friction to 0.05 reduced the slip to 0.002\,mm. Giving the pads the plate's constraint parameters ($\mathrm{solref}=[0.004, 1]$, $\mathrm{solimp}=[0.95, 0.99, 0.001]$), which contact priority had previously replaced with MuJoCo defaults, removed the remainder. No weld or attachment constraint is used.

\subsection{Workspace Departure of DrQ-v2+}
DrQ-v2+ scores 0\% on eight of the nine main tasks. In 20 evaluation episodes of one DrQ-v2+ checkpoint on \texttt{reach\_target\_single} (seeds $\seedbase$--$\seedbase+19$), the robot walked away from the workspace in every episode, backwards and sideways: the maximum pelvis displacement from the start ranged from 1.38 to 7.13\,m (mean 4.20\,m), and all 20 episodes timed out. These rollouts show how DrQ-v2+ fails; they do not isolate the lack of action chunking as the cause.

\FloatBarrier
\section{Coding-Agent Development and Execution Protocol}
\label{sec:app-agent}
\label{app:agent}

\subsection{Cold-Start Development Protocol and Sandbox}
We evaluate GPT-6 Astra (\texttt{gpt-6-astra}) in the official Codex CLI harness (v0.153.4) and Claude Opus 5.5 (\texttt{claude-opus-5-5}) in Claude Code (v2.1.280), both at high reasoning effort. The agent runs in a Docker container on an internal network whose only external route is a proxy allowlisted to the model provider's API. It receives:
\begin{enumerate}
  \item the task prompt (Listing~A1);
  \item the environment API documentation, including the observation and action layouts (Listing~A2);
  \item one human demonstration as three synchronised $84\times84$ MP4 files (\texttt{head}, \texttt{left\_wrist}, \texttt{right\_wrist}) with frame metadata.
\end{enumerate}

Each session has a budget of 101,000 environment steps, with 200 steps charged per reset. Only NumPy, SciPy, OpenCV, Pillow and the Python standard library are installed. The prompt forbids learned components, meaning networks, regressions or lookup tables fitted to data; this rule is stated in the prompt and is not enforced by an automatic check.

\subsection{Strict Interface versus Robotics Tooling}
We run 27 development sessions per interface, three independent sessions for each of the nine tasks.

\textbf{1. Strict interface (main comparison).} The agent has the same observations and action channels as the learned policies: three $84\times84$ RGB views, 50- or 56-dimensional proprioception, and absolute arm joint targets with body commands, without rate limiting. It has no camera calibration, depth, pixel-to-ray projection or inverse kinematics, and develops on the 60 demonstration seeds. Body commands are clipped at $\pm1.0$\,m/s and $\pm1.0$\,rad/s rather than the demonstration bounds that limit the learned policies (Table~\ref{tab:spaces}). In the logged evaluation episodes, 95\% of GPT-6 Astra's control steps stay within those bounds.

\textbf{2. Tool-augmented interface (reference).} The agent additionally receives $640\times480$ images, pinhole camera calibration, pixel-to-ray projection, world-frame wrist and base positions, Mink damped differential inverse kinematics and a command rate limiter (6\,rad/s with low-pass filtering), and develops on seeds 0--199.

Table~\ref{tab:agent-strict-tools} gives the per-session results, discussed in Section~\ref{sec:agent}. Listing~A3 is the complete strict-interface program for multi-modal reaching, from these pre-release sessions.

Table~\ref{tab:agent-usage} gives the per-session token use and API-equivalent cost of these pre-release strict-interface sessions; the most expensive single session cost \$31.23. The 27 tool-augmented sessions used 161.5\,M input tokens and \$235.07.

\begin{table*}[!htbp]
\centering\small
\caption{\textbf{Per-session success and cost of GPT-6 Astra under the strict and tool-augmented interfaces}, from the same sessions as Table~\ref{tab:agent-tools}. Each session's frozen program is evaluated on 100 hidden seeds; Mean $\pm$ SE is across the three sessions. Cost is the mean per-session API-equivalent cost. On block stacking, ACT reaches 0.13\% and tool-augmented Astra 6.3\% (three-session mean; best session 19\%).}
\label{tab:agent-strict-tools}
{\resizebox{\textwidth}{!}{\setlength{\tabcolsep}{2.5pt}\begin{tabular}{@{}l rrrr r rrrr r r@{}}
\toprule
 & \multicolumn{5}{c}{\textbf{Strict Interface} (84$\times$84 RGB, direct joint targets)} & \multicolumn{5}{c}{\textbf{Tool-Augmented Interface} ($640\times 480$, Calib, IK, DLS)} & \\
\cmidrule(lr){2-6} \cmidrule(lr){7-11}
\textbf{Task} & Run 1 & Run 2 & Run 3 & Mean\,$\pm$\,SE & Cost (\$) & Run 1 & Run 2 & Run 3 & Mean\,$\pm$\,SE & Cost (\$) & $\Delta$ \\
\midrule
Reach multi-modal       & 100 &  99 & 100 & 100\,$\pm$\,\phantom{0}0  & \phantom{0}3.2 & 100 & 100 & 100 & 100\,$\pm$\,\phantom{0}0  & \phantom{0}2.1 &   0 \\
Reach dual              &  98 &  91 &  94 &  94\,$\pm$\,\phantom{0}2  & \phantom{0}5.3 &  99 & 100 &  98 &  99\,$\pm$\,\phantom{0}1  & \phantom{0}5.2 &  +5 \\
Drawer close            & 100 & 100 & 100 & 100\,$\pm$\,\phantom{0}0  & \phantom{0}1.8 & 100 & 100 & 100 & 100\,$\pm$\,\phantom{0}0  & \phantom{0}2.3 &   0 \\
Drawer open             &   5 &  63 &  96 &  55\,$\pm$\,27            &           10.5 & 100 &  98 &  99 &  99\,$\pm$\,\phantom{0}1  & \phantom{0}5.2 & +44 \\
Move plate              &  31 &  40 &  47 &  39\,$\pm$\,\phantom{0}5  &           10.8 &  93 &  18 &  12 &  41\,$\pm$\,26            &           11.7 &  +2 \\
Move two plates         &   0 &   0 &   1 & \phantom{0}0$^*$\,$\pm$\,\phantom{0}0 & 25.3 &   4 &   0 &  78 &  27\,$\pm$\,25            &           15.4 & +27 \\
Pick box                &  63 &   8 &   0 &  24\,$\pm$\,20            &           18.2 &  55 &  11 &  37 &  34\,$\pm$\,13            & \phantom{0}9.9 & +11 \\
Dishwasher close        & 100 &   0 &   0 &  33\,$\pm$\,33            &           19.7 & 100 & 100 & 100 & 100\,$\pm$\,\phantom{0}0  &           16.2 & +67 \\
Dishwasher load cups    &   0 &  44 &  51 &  32\,$\pm$\,16            &           14.0 &  75 & 100 &  80 &  85\,$\pm$\,\phantom{0}8  &           10.4 & +53 \\
\midrule
\textbf{Mean}           & 55.2 & 49.4 & 54.3 & \textbf{53}             & \textbf{12.1}  & 80.7 & 69.7 & 78.2 & \textbf{76}             & \textbf{\phantom{0}8.7} & \textbf{+23} \\
\textbf{Total (27 sessions)} & \multicolumn{3}{c}{1431/2700 (53.00\%)} & --- & \textbf{326.3} & \multicolumn{3}{c}{2057/2700 (76.19\%)} & --- & \textbf{235.1} & --- \\
\midrule[\heavyrulewidth]
\multicolumn{12}{@{}l}{\footnotesize $^*$1 success in 300 episodes.} \\
\bottomrule
\end{tabular}
}}
\end{table*}

\begin{table}[!htbp]
\centering\small
\caption{\textbf{Per-session token use and cost of GPT-6 Astra under the strict interface} on the pre-release build, from the same sessions as Table~\ref{tab:agent-tools}. Three sessions per task.}
\label{tab:agent-usage}
{\setlength{\tabcolsep}{4pt}\begin{tabular}{@{}lrrrr@{}}
\toprule
Task & \multicolumn{3}{c}{API-equivalent cost (\$)} & Input tokens \\
\cmidrule(lr){2-4}
 & Run 1 & Run 2 & Run 3 & median (M) \\
\midrule
Reach multi-modal & 2.72 & 4.33 & 2.65 & 1.5 \\
Reach dual & 3.09 & 6.72 & 6.08 & 3.7 \\
Move plate & 11.19 & 11.36 & 9.71 & 8.4 \\
Move two plates & 22.87 & 31.23 & 21.78 & 20.7 \\
Dishwasher close & 13.22 & 21.63 & 24.17 & 16.6 \\
Dishwasher load cups & 15.08 & 12.68 & 14.35 & 10.2 \\
Drawer open & 15.02 & 6.66 & 9.77 & 6.5 \\
Drawer close & 1.32 & 2.21 & 1.76 & 1.0 \\
Pick box & 10.85 & 28.68 & 15.20 & 11.8 \\
\midrule
Total (27 sessions) & \multicolumn{3}{c}{326.33} & 237\phantom{.0} \\
\bottomrule
\end{tabular}
}
\end{table}

\subsection{Recalled Kinematics under the Strict Interface}
\label{app:agent-knowledge}
The strict interface provides no inverse kinematics, and the simulator's model files are not readable from the agent's workspace. Claude Opus 5.5 nonetheless modelled the G1 arms itself. In 22 of its 27 scored programs, the arm chain uses link offsets that match the original release of Unitree's public G1 description (\texttt{g1\_29dof.urdf}) to five significant figures, for example the shoulder-pitch joint at $(0.0039563, 0.10022, 0.23778)$\,m from the torso. All seven arm offsets in its forward kinematics match that description; the simulator uses Unitree's later revision with \mbox{Dex1-1} grippers, which differs in two of them (shoulder pitch at $z = 0.24778$\,m, wrist yaw at 0.051\,m rather than 0.046\,m), so the values were recalled from pretraining, not read from the simulator. In one session the agent's closing message states that ``the arm model uses G1 link lengths from memory of the robot description.'' Of the 27 programs, 24 solve inverse kinematics numerically, 18 of them with \texttt{scipy.optimize.least\_squares}; the three without are the drawer-closing sessions. No GPT-6 Astra program contains these values or solves inverse kinematics.

Listings~A3 and A4 show the two routes on the same task, multi-modal reaching. Astra's program thresholds the target colour in a wrist camera and moves the shoulder joints in proportion to its pixel offset. The Opus 5.5 program rebuilds the toolchain that the strict interface withholds: it models the head camera, lifts the target to 3D, walks to it and reaches it through forward kinematics from the recalled link offsets and a numerical IK. Both solve all 100 hidden seeds.

Table~\ref{tab:agent-sessions} lists every session. Opus 5.5 leads on multi-modal reaching and pick-box; Astra, without inverse kinematics, does better on dual reaching and dishwasher closing. Both agents fail most plate-transport and cup-loading episodes. The spread across sessions is large for both agents: a task is often solved in one session and failed in the others, so the three-session mean reflects how often a session finds a working strategy.

\begin{table}[!htbp]
\centering\small
\caption{\textbf{Per-session success (\%) of the coding agents under the strict interface}. Each session's frozen program is evaluated on 100 hidden seeds. $^\ast$: the scored program solves inverse kinematics.}
\label{tab:agent-sessions}
{\setlength{\tabcolsep}{4pt}\begin{tabular}{@{}lcccccc@{}}
\toprule
 & \multicolumn{3}{c}{Claude Opus 5.5} & \multicolumn{3}{c}{GPT-6 Astra} \\
\cmidrule(lr){2-4} \cmidrule(lr){5-7}
Task & s1 & s2 & s3 & s1 & s2 & s3 \\
\midrule
Pick box & 44$^\ast$ & 99$^\ast$ & 71$^\ast$ & 64 & 85 & 12 \\
Reach multi-modal & 100$^\ast$ & 100$^\ast$ & 100$^\ast$ & 75 & 97 & 95 \\
Reach dual & 71$^\ast$ & 99$^\ast$ & 65$^\ast$ & 100 & 93 & 75 \\
Drawer close & 100 & 100 & 100 & 96 & 100 & 100 \\
Drawer open & 99$^\ast$ & 98$^\ast$ & 100$^\ast$ & 100 & 97 & 100 \\
Move plate & 21$^\ast$ & 15$^\ast$ & 23$^\ast$ & 36 & 2 & 4 \\
Move two plates & 8$^\ast$ & 15$^\ast$ & 22$^\ast$ & 0 & 38 & 9 \\
Dishwasher close & 0$^\ast$ & 0$^\ast$ & 100$^\ast$ & 100 & 100 & 100 \\
Dishwasher load cups & 83$^\ast$ & 1$^\ast$ & 1$^\ast$ & 18 & 16 & 30 \\
\midrule
Programs with IK & \multicolumn{3}{c}{24 / 27} & \multicolumn{3}{c}{0 / 27} \\
\bottomrule
\end{tabular}
}
\end{table}

\FloatBarrier
\subsection{Agent Prompt, Task Sentences, API Document and an Example Program}
\label{sec:app-prompt-spec}
Listing~A1 is the prompt of every strict-interface session, verbatim; only its last line, the task sentence, differs between tasks (Table~\ref{tab:agent-task-sentences}). Tool-augmented sessions received the same template with $640\times480$ demonstration videos and training seeds 0--199. Listing~A2 is the strict-interface API document, verbatim for the 20-dimensional tasks. On 21-dimensional tasks it adds the torso-pitch action at index 4 and the waist joints at the front of the proprioceptive vector; the tool-augmented document additionally exposes inverse kinematics, camera calibration, pixel-to-ray projection and world-frame wrist and base positions. Entries 46--49 of the proprioceptive vector are the simulator's pelvis joint positions (Table~\ref{tab:spaces}). Listing~A3 is a complete submitted program.

\agentfile{boxtext}{Listing A1: Strict-interface prompt (verbatim text, line breaks reflowed; \{task\_sentence\} from Table~\ref{tab:agent-task-sentences})}{listings/prompt_strict.txt}

\par\medskip\noindent\begin{minipage}{\textwidth}\centering
\captionof{table}{\textbf{Task sentences} ending the agent prompt, identical under both interfaces. Agent sessions were run only on these tasks. The sentences were written for the agent and differ from the \pizero{} instructions (Table~\ref{tab:vla-instructions}). $^\dagger$Tool-augmented sessions only.}
\label{tab:agent-task-sentences}
\small
\begin{tabular}{@{}l>{\raggedright\arraybackslash}p{0.72\textwidth}@{}}
\toprule
Task & Task sentence \\
\midrule
Pick box & Pick the cardboard box up from the side table and gently set it down on the kitchen counter. \\
Reach multi-modal & Touch the red sphere with either hand. \\
Reach dual & Touch the red sphere with your left hand and the green sphere with your right hand at the same time. \\
Drawer close & Close the open top drawer of the kitchen cabinet. \\
Drawer open & Open the top drawer of the kitchen cabinet. \\
Move plate & Take the plate out of the left dish rack and gently stand it in the right dish rack. \\
Move two plates & Take both plates out of the left dish rack and gently stand them in the right dish rack. \\
Dishwasher close & Push the dishwasher racks in and close the door. \\
Dishwasher load cups & Gently place the two mugs from the counter into the upper rack of the dishwasher. \\
Stack blocks$^\dagger$ & Move the blocks to the other table and stack them on the green pad. \\
\bottomrule
\end{tabular}
\end{minipage}\par\medskip

\agentfile[boxDoc]{boxtext}{Listing A2: Strict-interface API document docs/api.md for 20-dimensional tasks (verbatim text, line breaks reflowed)}{listings/api_strict.md}

\agentfile[boxCode]{boxcode}{Listing A3: Complete policy.py written by GPT-6 Astra, strict interface, multi-modal reaching (100/100 hidden seeds; verbatim; from the same sessions as Table~\ref{tab:agent-tools})}{listings/policy_reach_multi_modal.py}

\agentfile[boxCode]{boxcode}{Listing A4: Excerpt of policy.py written by Claude Opus 5.5, strict interface, multi-modal reaching (100/100 hidden seeds; [...] marks elided lines; full program in the code release). The offsets in fk() are recalled from Unitree's g1\_29dof.urdf}{listings/policy_opus_reach_multi_modal_excerpt.py}

\end{document}